\newif\ifconfver

\newif\ifonecoltab

\newif\ifplainver  
\plainvertrue

\ifplainver
    \confverfalse                         
\fi

\ifconfver
     \documentclass[10pt,twocolumn,twoside]{IEEEtran}
\else
    \ifplainver
        \documentclass[11pt]{article}
        \usepackage{fullpage}
    \else
        \documentclass[12pt,draftcls,onecolumn]{IEEEtran}
    \fi
\fi

\usepackage{calc,amsfonts,amssymb,amsmath,bm,url,color,theorem,graphicx,cite}
\usepackage{psfrag,subfigure,float,epstopdf}
\usepackage[algoruled,linesnumbered]{algorithm2e}
\usepackage{multirow}
\usepackage[normalem]{ulem}

\definecolor{orange}{RGB}{255,107,0}

\newtheorem{Lemma}{Lemma}
\newtheorem{Prop}{Proposition}
\newtheorem{Theorem}{Theorem}
\newtheorem{Def}{Definition}

\newtheorem{Property}{Property}

\theorembodyfont{\rmfamily}

\newtheorem{Remark}{Remark}

\begin{document}

\bibliographystyle{IEEEtran}

\newcommand{\papertitle}{
Robust Volume Minimization-Based Matrix Factorization for Remote Sensing and Document Clustering
}

\newcommand{\paperabstract}{
This paper considers \emph{volume minimization} (VolMin)-based
structured matrix factorization (SMF).
VolMin is a factorization criterion that decomposes a given data matrix into a basis matrix times a structured coefficient matrix via finding the minimum-volume simplex that encloses all the columns of the data matrix.
Recent work showed that VolMin guarantees the identifiability of the factor matrices under mild conditions that are realistic in a wide variety of applications. This paper focuses on both theoretical and practical aspects of VolMin.
On the theory side, exact equivalence of two independently developed sufficient conditions for VolMin identifiability is proven here, thereby providing a more comprehensive understanding of this aspect of VolMin.
On the algorithm side, computational complexity and sensitivity to outliers are two key challenges associated with real-world applications of VolMin. These are addressed here via a new VolMin algorithm that handles volume regularization in a computationally simple way, and automatically detects and { iteratively downweights} outliers, simultaneously.
Simulations and real-data experiments using a remotely sensed hyperspectral image and the Reuters document corpus are employed to showcase the effectiveness of the proposed algorithm.
}


\ifplainver

    \date{\today}

    \title{\papertitle\footnote{Part of this work was published in IEEE ICASSP 2016, Shanghai, China \cite{fu2016icassp}.}}

    \author{
    $^\ast$Xiao Fu, $^\ast$Kejun Huang, $^\ast$Bo Yang,  $^\dag$Wing-Kin Ma, and $^\ast$Nicholas, D. Sidiropoulos
    \\ ~ \\
		$^\ast$Dept Elec. Computer Eng., University of Minnesota,\\
		Minneapolis, 55455, MN, United States\\
		Email: (xfu,huang663,yang4173,nikos)@umn.edu
	  \\~\\
    $^\dag$Dept Electronic Eng., The Chinese University of Hong  Kong, \\
    Shatin, N.T., Hong Kong \\
    Email: wkma@ieee.org
		\\
    }

    \maketitle

    \begin{abstract}
    \paperabstract
    \end{abstract}

\else
    \title{\papertitle}

    \ifconfver \else {\linespread{1.1} \rm \fi

   \author{Xiao Fu, \IEEEmembership{Member, IEEE}, Kejun Huang, \IEEEmembership{Student Member, IEEE}, Bo Yang, \IEEEmembership{Student Member, IEEE},  Wing-Kin Ma, \IEEEmembership{Senior Member, IEEE},  Nicholas D. Sidiropoulos$^\dag$\thanks{$^\dag$Corresponding Author.}, \IEEEmembership{Fellow, IEEE}
   
\thanks{This work was supported in part by the National Scientific Foundation (NSF) under Project NSF-ECCS 1608961 and in part by the Hong Kong Research Grants Council (RGC) under Project CUHK 14205414. A conference version of part of this work appears in {\em Proc. ICASSP 2016} \cite{fu2016icassp}.}
\thanks{X. Fu, K. Huang, Bo Yang, and N.D. Sidiropoulos are with the Department of Electrical and Computer Engineering, University of Minnesota, Minneapolis, MN55455, e-mail (xfu,huang663,yang4173,nikos)@umn.edu.}
\thanks{W.-K. Ma is with the Department of Electronic Engineering, the Chinese University of Hong Kong, Shatin, N.T., Hong Kong, e-mail wkma@ee.cuhk.edu.hk.
}
}

    \maketitle

    \ifconfver \else
        \begin{center} \vspace*{-2\baselineskip}
        \end{center}
    \fi

    \begin{abstract}
    \paperabstract
    \end{abstract}

    \begin{keywords}\vspace{-0.0cm}
       Simplex-volume minimization, identifiability, matrix factorization, robustness against outliers, hyperspectral unmixing, document clustering
    \end{keywords}

    \ifconfver \else \IEEEpeerreviewmaketitle} \fi

 \fi

\ifconfver \else
    \ifplainver \else
        \newpage
\fi \fi

\section{Introduction}
Structured matrix factorization (SMF) has been a popular tool in signal processing and machine learning.
For decades, factorization models such as the singular value decomposition (SVD) and eigen-decomposition have been applied for dimensionality reduction (DR), subspace estimation, noise suppression, feature extraction, etc.
Motivated by the influential paper of Lee and Seung \cite{lee1999learning}, new SMF models such as \emph{nonnegative matrix factorization} (NMF) have drawn much attention,
since they are capable of not only reducing dimensionality of the collected data,
but also retrieving loading factors that have physically meaningful interpretations.

In addition to NMF, some related SMF models have attracted considerable interest in recent years. The remote sensing community has spent much effort on a class of factorizations where the columns of one factor matrix are constrained to lie in the unit simplex \cite{Ma2013}. The same SMF model has also been utilized for document clustering \cite{gillis2014and},
and, most recently, multi-sensor array processing and blind separation of power spectra for dynamic spectrum access \cite{fu2015blind,fu2014sam}.

The first key question concerning SMF lies in identifiability -- when does a factorization model or criterion admit unique solution in terms of its factors?
Identifiability is
important in applications such as
parameter estimation, feature extraction, and signal separation. In recent years, identifiability conditions have been investigated for the NMF model \cite{donoho2003does,laurberg2008theorems,HuaSidSwa2014}.
An undesirable property of NMF highlighted in \cite{HuaSidSwa2014} is that identifiability hinges on both loading factors containing a certain number of zeros.
In many applications, however, there is at least one factor that is dense.
In hyperspectral unmixing (HU), for example, the basis factor (i.e., the spectral signature matrix) is always dense.
On the other hand, very recent work \cite{fu2015blind,lin2014identifiability} showed that the SMF model with the coefficient matrix columns lying in the unit simplex admits much more relaxed identifiability conditions.
Specifically, { Fu \emph{et al.} \cite{fu2015blind} and Lin \emph{et al.} \cite{lin2014identifiability} proved that}, under some realistic conditions, unique loading factors (up to column permutations) can be obtained by finding a minimum-volume enclosing
simplex of the data vectors.
Notably, these identifiability conditions of the so-called \emph{volume minimization} (VolMin) criterion allow working with dense basis matrix factors;
in fact, the model does not impose any constraints on the basis matrix except for having full-column rank.
Since the NMF model can be recast as (viewed as a special case of) the above SMF model \cite{Gillis2012},
such results suggest that VolMin is an attractive alternative to NMF for the wide range of applications of NMF and beyond.

Compared to NMF, VolMin-based matrix factorization is computationally more challenging.
The notable prior works
in \cite{li2008minimum} and \cite{MVES} formulated VolMin as a constrained (log-)determinant minimization problem, and applied successive convex optimization and
alternating optimization to deal with it, respectively. The major drawback of these pioneering works is that the algorithms were developed under a noiseless setting, and thus only work well
for high signal-to-noise ratio (SNR) cases.
Also,
these algorithms work in the dimension-reduced domain, but the DR process may be sensitive to outliers and modeling errors.
The work \cite{ambikapathi2011chance} took noise into consideration, but the algorithm is computationally prohibitive and has no guarantee of convergence.
Some other algorithms \cite{miao2007endmember,zhou2011minimum} work in the original data domain, and deal with a
volume-regularized
data fitting problem.
Such a formulation can tolerate noise to a certain level,
but is harder to tackle than those in \cite{li2008minimum,MVES,zhou2011minimum} -- volume regularizers typically introduce extra difficulty to an already very hard bilinear fitting problem.

The second major challenge of implementing VolMin is that
the VolMin
criterion is very sensitive to outliers:
it has been noted in the literature that even a single outlier can make the VolMin criterion fail \cite{Ma2013}.
However, in real-world applications, outlying measurements are commonly seen:
in HU,
pixels that do not
always
obey the
nominal
model are frequently spotted because of the complicated physical environment \cite{dobigeon2014nonlinear}; and in document clustering, articles that are difficult to be classified to any known category may also act like outliers.
The algorithm in \cite{bioucas2009variable} is the state-of-the-art VolMin algorithm that takes outliers into consideration.
It imposes a `soft penalty' on outliers that lie outside the simplex that is sought, thereby allowing the existence of some outliers and achieving robustness.
The algorithm works fairly well when the data are not severely corrupted, but it works in the reduced-dimension domain -- and DR pre-processing can fail due to outliers.

\bigskip

\noindent
{\bf Contributions}
In this work, we explore both theoretical and practical aspects of VolMin.
On the theory side, we show that two existing sufficient conditions for VolMin identifiability are in fact equivalent.
The two identifiability results were developed in parallel, rely on different mathematical tools, and offer seemingly different characterizations of the sufficient conditions -- so their equivalence is not obvious. Our proof `cross-validates' the existing results, and thus leads to a deeper understanding of the VolMin problem.

On the algorithm side, we propose a new algorithmic framework for dealing with the VolMin criterion.
The proposed framework takes outliers into consideration, without requiring DR pre-processing.
Specifically,
we impose an outlier-robust loss function onto the data fitting part, and propose a modified log-determinant loss function as the volume regularizer.
By majorizing both functions, the fitting and the volume-regularization terms can be taken care of in a refreshingly easy way, and a simple \emph{inexact} alternating optimization algorithm is derived.
A Nesterov-type first-order optimization technique is further employed within this framework
to accelerate convergence.
The proposed algorithm is flexible -- problem-specific prior information on the factors and different volume regularizers can be easily incorporated. Convergence of the proposed algorithm to a stationary point is also shown.

Besides a judiciously designed set of simulations,
we also validate the proposed algorithm using real-life datasets.
Specifically, we use remotely sensed hyperspectral image data and document data
to showcase the effectiveness of the proposed algorithm in hyperspectral unmixing and document clustering applications, respectively.
Notice that VolMin has never been used for document clustering before, to the best of our knowledge,
and our work shows that VolMin is indeed very effective in this context, outperforming the state-of-art in terms of clustering accuracy.

A conference version of part of this work appears in \cite{fu2016icassp}.
Beyond \cite{fu2016icassp}, this journal version includes the equivalence of the identifiability conditions,
first-order optimization-based updates, consideration of different types of regularization and constraints,
proof of convergence, extensive simulations, and experiments using real data.


%
%



~\\
\emph{Notation}:
We largely follow common notational conventions in signal processing.
${\bm x} \in \mathbb{R}^n$ and ${\bm X} \in \mathbb{R}^{m \times n}$ denote a real-valued $n$-dimensional vector and a real-valued $m \times n$ matrix, respectively (resp.).
${\bm x} \geq {\bm 0}$ (resp. ${\bm X} \geq {\bm 0}$) means that ${\bm x}$ (resp. ${\bm X}$) is element-wise non-negative.
${\bm x}\in\mathbb{R}^n_+$ (resp. ${\bm X} \in \mathbb{R}^{m \times n}_+$) also means that
${\bm x}$ (resp. ${\bm X}$) is element-wise non-negative.
${\bm X}\succ {\bm 0}$ and ${\bm X}\succeq {\bm 0}$ mean that ${\bm X}$ is positive definite and positive semidefinite, resp.
The superscripts ``$T$'' and ``$-1$'' stand for the transpose and inverse operations, resp.
The $\ell_p$ norm of a vector ${\bm x}\in\mathbb{R}^n$, $p \geq 1$, is denoted by $\| {\bm x}\|_p = (\sum_{i=1}^n |x_i|^p)^{1/p}$.
The $\ell_p$ quasi-norm, $0 < p < 1$, is denoted by the same notation.
The Frobenious norm and the matrix 2-norm are denoted by $\|{\bm X}\|_F$ and $\| {\bm X} \|_{2}$, respectively.
The all-one vector is denoted by ${\bm 1}$.

In this paper, we also make extensive use of convex analysis.
Let ${\bm X}=[{\bm x}_1,\ldots,{\bm x}_m]$.
The convex cone of ${\bm x}_1,\ldots,{\bm x}_m$ is denoted by ${\rm cone}\{{\bm x}_1,\ldots,{\bm x}_m\}={\rm cone}({\bm X})=\{{\bm y}~|~{\bm y}={\bm X}{\bm \theta},~{\bm \theta}\geq{\bm 0}\}$;
the convex hull of ${\bm x}_1,\ldots,{\bm x}_m$ is denoted by ${\rm conv}\{{\bm x}_1,\ldots,{\bm x}_m\} = {\rm conv}({\bm X})=\{{\bm y}~|~{\bm y}={\bm X}{\bm \theta},~{\bm \theta}\geq{\bm 0},{\bm 1}^T{\bm \theta}=1\}$;
when $\{{\bm x}_1,\ldots,{\bm x}_m\}$ are linearly independent, ${\rm conv}({\bm X})$ is also called a simplex;
the set of extreme rays of ${\rm cone}({\bm X})$ is denoted by ${\rm ex}\{{\rm cone}({\bm X})\}$;
and the dual cone of a convex ${\cal X}$ is denoted by ${\cal X}^\ast = \{{\bm y}~|~{\bm y}^T{\bm x}\geq{\bm 0},~{\bm x}\in{\cal X}\}$;
${\rm bd}{\cal X}$ denotes the set of the boundary points of the \emph{second order cone} ${\cal X}$.
We point the readers to \cite{CVX,fu2015blind,HuaSidSwa2014} for detailed illustration of the above concepts.

\section{The VolMin Criterion and Identifiability}
%
%
In this section, we first give a brief introduction to the VolMin criterion for SMF and a concise review of the existing identifiability results.
Then, we prove that the two independently developed identifiability results (using rather different mathematical tools) are equivalent.

\subsection{Background}
\begin{figure}
	\centering
	\psfrag{Ab}{\tiny ${\bm A}$}
	\psfrag{Xb}{\tiny ${\bm X}$}
	\psfrag{Sb}{\tiny ${\bm S}$}
	\psfrag{x[l] hyperspectral pixel (document)}{\footnotesize  ${\bm x}[\ell]$: a hyperspectral pixel / a document}
	\psfrag{ak material spectrum (topic)}{\footnotesize  ${\bm a}_k$: spectral signature of material $k$ / topic $k$.}	
	\psfrag{s[l] abundance }{\footnotesize $\begin{array}{l} \text{${\bm s}[\ell]$: proportion of materials (topics)}\\ \text{ in pixel (document) $\ell$ }\end{array}$}
		\includegraphics[width=7cm]{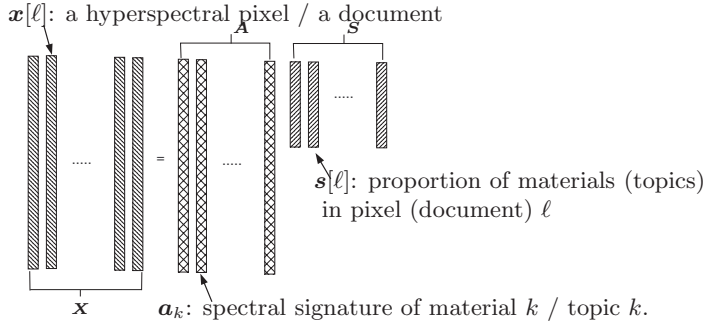}
	\caption{Motivating examples: Hyperspectral unmixing and document clustering.}
	\label{fig:motivation_rvolmin}
\end{figure}
Consider the following signal model:
\begin{equation}\label{eq:basic_mod}
	          {\bm x}[\ell] = {\bm A}{\bm s}[\ell]+{\bm v}[\ell],\quad \ell=1,\ldots,L,
\end{equation}
where $ {\bm x}[\ell] \in\mathbb{R}^{M}$ is a measured data vector that is indexed by $\ell$,
${\bm A}\in\mathbb{R}^{M\times K}$ is a basis which is assumed to have full column-rank,
${\bm s}[\ell]\in\mathbb{R}^{K}$ is the coefficient vector representing ${\bm x}[\ell]$ in the low dimensional subspace ${\rm range}({\bm A})$,
and ${\bm v}[\ell]\in\mathbb{R}^M$ denotes noise.
We assume that every ${\bm s}[\ell]$
satisfies
\begin{equation}\label{eq:sumtoone}
	{\bm s}[\ell]\geq{\bm 0}~\text{and}~{\bm 1}^T{\bm s}[\ell]=1.
\end{equation}
The model can be compactly written as ${\bm X}={\bm A}{\bm S}+{\bm V}$, where ${\bm X}=[{\bm x}[1],\ldots,{\bm x}[L]]$, ${\bm S}=[{\bm s}[1],\ldots,{\bm s}[L]]$ and ${\bm V}=[{\bm v}[1],\ldots,{\bm v}[L]]$.

The task of SMF is to factor ${\bm X}$ into ${\bm A}$ and ${\bm S}$.
The simple model in \eqref{eq:basic_mod}-\eqref{eq:sumtoone} parsimoniously captures the essence of a large variety of applications.
For \emph{document clustering} or \emph{topic mining} \cite{gillis2014and}, estimating ${\bm A}$ and ${\bm S}$ can help recognize the most popular topics/opinions in textual data (e.g., documents, web content, or social network posts),
and cluster the data according to their weights on different topics/opinions.
In \emph{hyperspectral} {\em remote sensing} \cite{fu2015self,Ma2013},
${\bm x}[\ell]$ represents a remotely sensed pixel using sensors of high spectral resolution,
${\bm a}_1,\ldots,{\bm a}_K$ denote $K$ different spectral signatures of materials that comprise the pixel ${\bm x}[\ell]$,
and ${s}_k[\ell]$ denotes the proportion of material $k$ contained in pixel ${\bm x}[\ell]$.
Estimating ${\bm A}$ enables recognition of the underlying materials in a hyperspectral image.
See Fig.~\ref{fig:motivation_rvolmin} for an illustration of these motivating examples.
Very recently, the same model has been applied to {\emph{power spectra separation}} \cite{fu2014sam} for dynamic spectrum access and \emph{fast blind speech separation} \cite{fu2015blind}.
In addition, many applications of NMF can also be considered under the model in \eqref{eq:basic_mod}-\eqref{eq:sumtoone}, after suitable normalization \cite{Gillis2012}.

\begin{figure}[!h]
\begin{center}
\psfrag{a1}{$\bm a_1$}
\psfrag{a2}{$\bm a_2$}
\psfrag{a3}{$\bm a_3$}
\psfrag{b1}{$\bm b_1$}
\psfrag{b2}{$\bm b_2$}
\psfrag{b3}{$\bm b_3$}
\psfrag{x}{${\bm x}[\ell]$}
\psfrag{0}{${\bm 0}$}
\psfrag{enclose}{\footnotesize  enclosing convex hull}
\psfrag{min}{\footnotesize  min.-vol. convex hull}
\includegraphics[width=5.5cm]{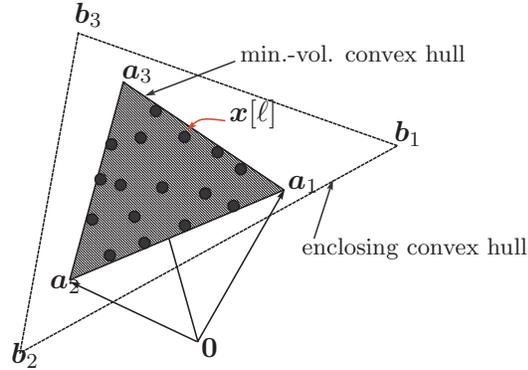}
\caption{\small The intuition of VolMin.} 
\label{fig:experiments}
\end{center}
\end{figure}

Many algorithms have been developed for finding such a factorization, and we refer the readers to \cite{Ma2013,gillis2014and} for a survey.
Among these algorithms, we are particularly interested in the so-called \emph{volume minimization}
(VolMin) criterion, which is identifiable under certain reasonable conditions.
VolMin is motivated by the nice geometrical interpretation of the constraints in \eqref{eq:sumtoone}: Under these constraints, all the data points live in a \emph{convex hull} spanned by ${\bm a}_1,\ldots,{\bm a}_K$ (or, a \emph{simplex} spanned by ${\bm a}_1,\ldots,{\bm a}_K$ when the ${\bm a}_k$'s are linearly independent); see Fig.~\ref{fig:experiments}.
If the data points are sufficiently spread in ${\rm conv}\{{\bm a}_1,\ldots,{\bm a}_K\}$,
then the minimum-volume enclosing convex hull coincides with ${\rm conv}\{{\bm a}_1,\ldots,{\bm a}_K\}$.
Formally, the VolMin criterion can be
formulated as
\begin{subequations}\label{eq:VolMin_form}
\begin{align}
	           ({\bm A},\{{\bm s}[\ell]\})=&\arg\min_{{\bm B},~\{{\bm c}[\ell]\}}~{\rm vol}({\bm B})\\
							     {\rm s.t.}~&{\bm x}[\ell] = {\bm B}{\bm c}[\ell],  \label{eq:simp1}\\
									 &{\bm 1}^T{\bm c}[\ell]=1,~{\bm c}[\ell]\geq{\bm 0},~\forall \ell,\label{eq:simp2}
\end{align}
\end{subequations}
where ${\rm vol}({\bm B})$ denotes a measure that is related or proportional to the volume of the simplex
${\rm conv}\{{\bm b}_1,\ldots,{\bm b}_K\}$, and
\eqref{eq:simp1}-\eqref{eq:simp2} mean that every ${\bm x}[\ell]$ is enclosed in
${\rm conv}\{{\bm b}_1,\ldots,{\bm b}_K\}$
(i.e., ${\bm x}[\ell] \in{\rm conv}\{{\bm b}_1,\ldots,{\bm b}_K\}$).
In the literature, various functions for ${\rm vol}({\bm B})$ have been used \cite{fu2015blind,bioucas2009variable,miao2007endmember,MVES,ambikapathi2011chance,li2008minimum,zhou2011minimum}.
One representative choice of ${\rm vol}({\bm B})$ is
\begin{equation} \label{eq:vol_a}
{\rm vol}({\bm B}) = {\rm det}( \bar{\bm B}^T \bar{\bm B} ), ~~
\bar{\bm B} = [~ \bm b_1 - \bm b_K, \ldots, \bm b_{K-1} - \bm b_K ~],
\end{equation}
or its variants; see \cite{MVES,ambikapathi2011chance,miao2007endmember}.
The reason of employing such a function is that $\sqrt{ {\rm det}( \bar{\bm B}^T \bar{\bm B} ) }/ ((N-1)!)$ is the volume of the simplex ${\rm conv}\{{\bm b}_1,\ldots,{\bm b}_K\}$ by definition \cite{gritzmann1995largestj}.
Another popular choice of ${\rm vol}({\bm B})$ is
\begin{equation} \label{eq:vol_b}
{\rm vol}({\bm B}) = {\rm det}( {\bm B}^T {\bm B} );
\end{equation}
see \cite{fu2015blind,bioucas2009variable,li2008minimum,zhou2011minimum}.
Note that $\sqrt{ {\rm det}( {\bm B}^T {\bm B} ) }/ (N!)$ is the volume of the simplex ${\rm conv}\{{\bm 0}, {\bm b}_1,\ldots,{\bm b}_K\}$,
which should scale similarly with the volume of ${\rm conv}\{{\bm b}_1,\ldots,{\bm b}_K\}$.
The upshot of \eqref{eq:vol_b} is that \eqref{eq:vol_b}
has a simpler structure than \eqref{eq:vol_a}.

\subsection{Identifiability of VolMin}
The most appealing aspect of VolMin is its identifiability of ${\bm A}$ and ${\bm S}$: Under mild and realistic conditions, the optimal solution to Problem~\eqref{eq:VolMin_form}
is essentially the true $({\bm A},{\bm S})$.
To be precise, let us make the following definition.
\begin{Def}(VolMin Identifiability)
Consider the matrix factorization model in \eqref{eq:basic_mod}-\eqref{eq:sumtoone}, and let $({\bm B}^\star,{\bm C}^\star )$ be any optimal solution to Problem~\eqref{eq:VolMin_form}.
If every optimal $({\bm B}^\star,{\bm C}^\star )$ satisfies ${\bm B}^\star={\bm A}{\bm \Pi}$ and ${\bm C}^\star=\bm \Pi^T{\bm S}$, where ${\bm \Pi}$ denotes a permutation matrix, then we say that VolMin identifies the true matrix factors, or \emph{VolMin identifiability holds}.
\end{Def}
{ Fu \emph{et al.} \cite{fu2015blind} have shown that}
\begin{Theorem}\label{thm:FuMaHuaSid}
Let ${\rm vol}(\bm B)$ be the function in \eqref{eq:vol_b}.
Define a second order cone ${\cal C}=\{{\bm x}\in\mathbb{R}^N~|~{\bm 1}^T{\bm x}\geq\frac{1}{\sqrt{N-1}}\|{\bm x}\|_2\}$.
Then VolMin identifiability holds if ${\rm rank}({\bm A})={\rm rank}({\bm S})=K$ and
\begin{enumerate}
		\item[i)] ${\cal C}\subseteq {\rm cone}({\bm S})$; and
		\item[ii)] ${\rm cone}({\bm S})\not\subseteq{\rm cone}({\bm Q})$ where ${\bm Q}$ is any unitary matrix except the permutation matrices.
	\end{enumerate}	
\end{Theorem}
In plain words, a sufficient condition under which VolMin identifiability holds is when $\{{\bm s}[\ell]\}_{\ell=1}^L$ are sufficiently scattered over the unit simplex, such that the second order cone ${\cal C}$
is a subset of ${\rm cone}({\bm S})$.
By comparing Theorem~\ref{thm:FuMaHuaSid} to the identifiability conditions for NMF (see \cite{donoho2003does,laurberg2008theorems,HuaSidSwa2014}),
we see a remarkable advantage -- VolMin does not have any restriction on ${\bm A}$ except being full-column rank.
In fact, ${\bm A}$ can be dense, partially negative, or even complex-valued. This result allows us to apply VolMin to a wider variety of applications than NMF.

\begin{figure}[!h]
\begin{center}
\psfrag{e1}{$\bm e_1$}
\psfrag{e2}{$\bm e_2$}
\psfrag{e3}{$\bm e_3$}
\psfrag{cone}{${\cal C}$ or ${\cal R}\left(\frac{1}{\sqrt{N-1}}\right)$}
\psfrag{1x=1}{${\bm 1}^T{\bm x}=1$.}
\includegraphics[width=5cm]{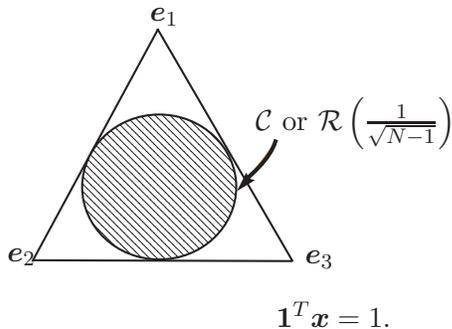}
\caption{\small Visualization of the sufficient conditions on the hyperplane ${\bm 1}^T{\bm x}=1$. The sufficient conditions in \cite{fu2015blind} and \cite{lin2014identifiability} both
require that ${\cal C}$ (the inner circle) is contained in ${\rm cone}({\bm S})$.}
\label{fig:suff}
\end{center}
\end{figure}
In \cite{lin2014identifiability},
another sufficient condition for VolMin identifiability was proposed:
\begin{Theorem}\label{thm:LinMa}
Let ${\rm vol}(\bm B)$ be the function in \eqref{eq:vol_a}.
Assume ${\rm rank}({\bm A})={\rm rank}({\bm S})=K$. Define ${\cal R}(r)=\{{\bm s}\in\mathbb{R}^N~|~\{ \|{\bm s}\|_2\leq r \} \cap {\rm conv}\{{\bm e}_1,\ldots,{\bm e}_N\} \}$ and $\gamma={\rm sup}\{ r~|~{\cal R}(r)\} \subseteq {\rm conv}({\bm S})\}$. Then VolMin identifiability holds if
	$\gamma>\frac{1}{\sqrt{N-1}}$.
\end{Theorem}

Theorem~\ref{thm:LinMa} does not characterize its identifiability condition using convex cones like Theorem~\ref{thm:FuMaHuaSid} did.
Instead, it defines a `diameter' $r$ of the convex hull spanned by the columns of ${\bm S}$,
and then develops an identifiability condition based on it.

The sufficient conditions presented in the two theorems seemingly have different flavors, but we notice that they are related in essence.
To see the connections,
we first note that Theorem~\ref{thm:FuMaHuaSid} still holds after replacing ${\rm cone}({\bm S})$ and ${\cal C}$ with convex hulls ${\rm conv}\{{\bm s}[1],\ldots,{\bm s}[L]\}$ and ${\cal C}\cap {\rm conv}\{{\bm e}_1,\ldots,{\bm e}_N\}$, respectively -- since the ${\bm s}[\ell]$'s are all in $ {\rm conv}\{{\bm e}_1,\ldots,{\bm e}_N\}$.
In fact, ${\cal C}\cap {\rm conv}\{{\bm e}_1,\ldots,{\bm e}_N\}$ is exactly the set ${\cal R}(r)$ for $r\leq 1/\sqrt{N-1}$.
Geometrically, we illustrate the conditions in Fig.~\ref{fig:suff} using $N=3$ for visualization.
We see that, if we look at the conditions in Theorem~\ref{thm:FuMaHuaSid} at the 2-dimensional hyperplane that contains
${\bm 1}^T{\bm x}=1$, the two conditions both mean that the inner shaded region is contained in ${\rm conv}\{{\bm s}[1],\ldots,{\bm s}[L]\}$. Motivated by this observation, in this paper, we rigorously show that

\begin{Theorem}\label{thm:equivalence}
	The sufficient conditions for VolMin identifiability in Theorem~\ref{thm:FuMaHuaSid} and Theorem~\ref{thm:LinMa} are equivalent.
\end{Theorem}
The proof of Theorem~\ref{thm:equivalence} can be found in Appendix~\ref{app:thm}.
Although the geometrical connection may seem clear on hindsight, rigorous proof is highly nontrivial.
We first show that the condition in Theorem~\ref{thm:FuMaHuaSid}
is equivalent to another condition,
and then establish equivalence between the `intermediate' condition
and the condition in Theorem~\ref{thm:LinMa}.

\begin{Remark}
The
equivalence between the sufficient conditions in Theorem~\ref{thm:FuMaHuaSid} and Theorem~\ref{thm:LinMa} is interesting and surprising -- although
the corresponding theoretical developments started from very different points of view, they converged to equivalent conditions.
Their equivalence brings us deeper understanding of the VolMin criterion.
The proof itself clarifies the role of regularity condition ii) in Theorem~\ref{thm:FuMaHuaSid}, which was originally difficult to describe geometrically -- and now we understand that condition ii) is there to ensure $\gamma>\frac{1}{\sqrt{N-1}}$, i.e., the existence of a convex cone that is `sandwiched' by ${\rm cone}({\bm S})$ and ${\cal C}$.
In addition, the equivalence also suggests that the different cost functions in \eqref{eq:vol_a} and \eqref{eq:vol_b}
ensure identifiability of ${\bm A}$ and ${\bm S}$ under the same sufficient conditions, and thus
they are expected to perform similarly in practice. On the other hand, since the function in \eqref{eq:vol_b} is easier to handle,
using it
in practice is more appealing.
As a by-product, since we have proved that condition ii) is equivalent to a condition that was used for NMF identifiability in \cite{HuaSidSwa2014} (cf. Lemma~\ref{lem:conditions}), our result here also helps better understand the sufficient condition for NMF identifibility in \cite{HuaSidSwa2014} in a more intuitively pleasing way.
\end{Remark}

\section{Robust VolMin via Inexact BCD}
In this section, we turn our attention to designing algorithms for dealing with the VolMin criterion.
Optimizing the VolMin criterion is challenging.
In early works such as \cite{li2008minimum,bioucas2009variable},
linear DR with $\bm X$ is assumed such that the basis after DR is a square matrix.
This subsequently enables one to write the DR-domain VolMin problem as
\begin{equation}\label{eq:MVSA}
\begin{aligned}
\min_{\substack{ \tilde{\bm B} \in \mathbb{R}^{K\times K},   {\bm C} \in \mathbb{R}^{K \times L} }}~&\log | \det(\tilde{\bm B}) | \\
{\rm s.t.}~& \tilde{\bm x}[\ell]=\tilde{\bm B}{\bm c}[\ell]\\
           &{\bm 1}^T{\bm c}[\ell]=1,~{\bm c}[\ell]\geq {\bm 0},
\end{aligned}
\end{equation}
where
$\tilde{\bm x}[\ell]\in\mathbb{R}^K$ is the dimension-reduced data vector corresponding to $\bm x[\ell]$, and $\tilde{\bm B}\in\mathbb{R}^{K\times K}$ is a dimension-reduced basis.
Note that minimizing $\log | \det(\tilde{\bm B}) |$ is the same as minimizing $\det(\tilde{\bm B}^T \tilde{\bm B})$.
Problem~\eqref{eq:MVSA} can be efficiently tackled via either alternating optimization \cite{MVES} or successive convex optimization \cite{li2008minimum,bioucas2009variable}.
The drawback with these existing
algorithms is that noise was not taken into consideration.
Also, these approaches require DR to make the effective ${\bm A}$ square -- but DR may not be reliable in the presence of outliers or modeling errors.
Another major class of algorithms such as those in \cite{miao2007endmember,zhou2011minimum} considers 
\begin{equation} \label{eq:reg}
    \begin{aligned}
     \min_{\substack{ {\bm B} \in \mathbb{R}^{M\times K},   {\bm C} \in \mathbb{R}^{K \times L} }}  & ~\left\|{\bm X}-{\bm B}{\bm C}\right\|_F^2 + \lambda\cdot{\rm vol}({\bm B})\\
                {\rm s.t.}      & ~  {\bm C}\geq {\bf 0},~{\bf 1}^T {\bm C}={\bm 1}^T,
    \end{aligned}
\end{equation}
where
$\lambda>0$ is a parameter that balances data fidelity versus volume minimization.
The formulation in \eqref{eq:reg} avoids DR and takes noise into consideration.
However, our experience is that volume regularizers, such as  ${\rm vol}(\bm B)= \det({\bm B}^T{\bm B})$, are numerically harder to cope with, which will be explained in detail later.
{ 
We should also compare Problem~\eqref{eq:reg} with the VolMin formulation in \eqref{eq:VolMin_form}.
Problem~\eqref{eq:VolMin_form} enforces a hard constraint ${\bm X}={\bm B}{\bm C}$, and thus ensures that every feasible ${\bm B}$ and ${\bm C}$ have full rank in the noiseless case.
On the other hand, Problem~\eqref{eq:reg} employs a fitting-based criterion, and an overly large $\lambda$
could result in rank-deficient factors even in the noiseless case.
Hence, $\lambda$ should be chosen with caution.}

\begin{figure}[!t]
	\centering
	\psfrag{a1}{${\bm a}_1$}
	\psfrag{a2}{${\bm a}_2$}
	\psfrag{a3}{${\bm a}_3$}
		\includegraphics[width=6cm]{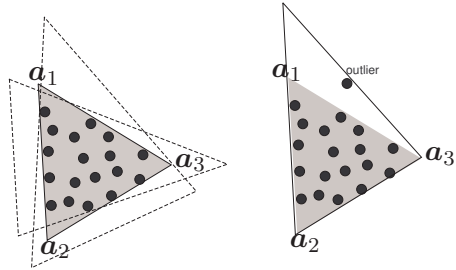}
	\caption{The impact of outliers to VolMin.
	The dots are ${\bm x}[\ell]$'s; the shaded area is ${\rm conv}\{{\bm a}_1,\ldots,{\bm a}_N\}$, the triangles with dashed lines are data-enclosing convex hulls, and the one with solid lines is the minimum-volume enclosing convex hull. Left: the case where no outliers exist.	Right: the case where a single outlier exists.}
	\label{fig:MinVol}
\end{figure}

Another notable difficulty is that outliers are very damaging to the VolMin criterion.
In many cases, a single outlier can make the minimum-volume enclosing convex hull very different from the desired one; see Fig.~\ref{fig:MinVol}
for an illustration.
The state-of-the-art algorithm that considers outliers for the VolMin-based factorization is
 \emph{simplex identification via split augmented Lagrangian} (SISAL)  \cite{bioucas2009variable}, but it takes care of outliers in the dimension-reduced domain.
As already mentioned,
the DR process itself may be impaired by outliers,
and thus dealing with outliers in the original data domain is more appealing.
Directly factoring ${\bm X}$ in the original data domain also has the advantage of allowing us to incorporate any \emph{a priori} information on ${\bm A}$ and ${\bm S}$, such as nonnegativity, smoothness, and sparsity.

\subsection{Proposed Robust VolMin Algorithm}
We are interested in the VolMin-regularized matrix factorization, but we take the outlier problem into consideration.
Specifically, we propose to employ the following optimization surrogate of the VolMin criterion:
\begin{align}\label{eq:R_VolMin}
	                \min_{{\bm B},{\bm C}}&~\sum_{\ell=1}^L\frac{1}{2}\left(\left\|{\bm x}[\ell]-{\bm B}{\bm c}[\ell] \right\|_2^2+\epsilon\right)^{\frac{p}{2}} +\frac{\lambda}{2}  \log\det({\bm B}^T{\bm B}+\tau{\bm I}) \nonumber\\
								  {\rm s.t.}&~{\bm 1}^T{\bm c}[\ell]=1,~{\bm c}[\ell]\geq{\bm 0},~\forall \ell,
\end{align}
where $p\in(0,2]$, $\lambda>0$,
$\epsilon > 0$, and $\tau > 0$.
Here, $\epsilon>0$ is a small regularization parameter,
which keeps the first term inside its smooth region for computational convenience when $p<1$;
if $p\in(1,2]$, we can simply let $\epsilon=0$.
The parameter $\tau>0$ is also a small positive number, which is used to ensure that the cost function is
bounded from below for any $\bm B$.

{ The motivation of using $\log\det({\bm B}^T{\bm B}+\tau{\bm I})$ instead of the commonly used volume regularizers such as $\det({\bm B}^T{\bm B})$ is computational simplicity: Although both functions are non-convex and conceptually equally hard to deal with, the former features a much simpler update rule because it admits a tight upper bound while the latter does not -- this point will become clearer shortly. Interestingly, $\log\det({\bm B}^T{\bm B}+\tau{\bm I})$ has been used in the context
of low-rank matrix recovery \cite{fazel2003log,liu2013robust}, but here we instead apply it for simplex-volume minimization.
The $\ell_2/\ell_p$-(quasi-) norm data fitting part is employed to downweight the impact of the outliers --  when $0<p< 2$, such a fitting criterion is
less sensitive to large fitting errors and thus is robust against outliers.
Other robust fitting criteria can also be considered -- e.g., the $\ell_p$ norm-based criterion $\|{\bm X}-{\bm B}{\bm C}\|_p^p$ for $0<p<2$ where $\|{\bm Y}\|_p^p=\sum_{i=1}^m\sum_{j=1}^m|Y_{i,j}|^p$ is known to be robust to entry-level outliers \cite{liu2015smoothing,ekblom1974p,RobustParafac}.
Nevertheless, the type of outliers that matters in VolMin is column outliers (or gross outliers) which represents a point lying outside the ground-truth convex hull, and the proposed criterion is natural for fending against such outliers.
In addition, computationally, the $\ell_2/\ell_p$ mixed-norm criterion can be handled efficiently, as we will see.
}

Our primary objective is to handle Problem~\eqref{eq:R_VolMin} efficiently.
Nonetheless, we will also show that the proposed algorithmic framework can easily incorporate different volume-associated regularizers
in the literature,
such as the previously mentioned ${\rm vol}({\bm B})=\det({\bm B}^T{\bm B})$,
and
\begin{equation}\label{eq:trace_vol}
	{\rm vol}({\bm B})=\sum_{i=1}^{K-1}\sum_{j=i+1}^K\|{\bm b}_i-{\bm b}_j\|_2^2;
\end{equation}
see \cite{berman2004ice}.
Notice that \eqref{eq:trace_vol} is a coarse approximation of
the volume of ${\rm conv}\{ \bm b_1,\ldots, \bm b_K \}$,
which measures the volume by
simply adding up the squared distances between the vertices.


\subsection{Update of ${\bm C}$}\label{sec:C}
Our idea is to update ${\bm B}$ and ${\bm C}$ alternately, i.e., using block coordinate descent (BCD).
Unlike classic BCD \cite{bertsekas1999nonlinear}, we solve the partial optimization problems in an \emph{inexact} fashion for efficiency.
We first consider updating ${\bm C}$.
The problem w.r.t. ${\bm C}$ is separable w.r.t. $\ell$ and convex.
Therefore, after $t$ iterations with the current solution $({\bm B}^t,{\bm C}^t)$, we consider:
\begin{equation}\label{eq:Cupdate}
\begin{aligned}
	                {\bm c}^{t+1}[\ell]:= \arg\min_{{\bm c}[\ell]}&\quad\frac{1}{2}\left\|{\bm x}[\ell]-{\bm B}^{t}{\bm c}[\ell]\right\|_2^2\\
									 {\rm s.t.}&\quad {\bm 1}^T{\bm c}[\ell]=1,~{\bm c}[\ell]\geq {\bm 0},								 
\end{aligned}
\end{equation}
for $\ell=1,\ldots,L$.
Since Problem~\eqref{eq:Cupdate} is convex, one can update ${\bm C}$ by solving Problem~\eqref{eq:Cupdate} to optimality.
An \emph{alternating direction method of multipliers} (ADMM)-based algorithm was provided in the conference version of this work for this purpose; see the detailed implementation in \cite{fu2016icassp}.
Nevertheless, exactly solving Problem~\eqref{eq:Cupdate} at each iteration is computationally costly, especially when the problem size is large.
Here, we propose to deal with Problem~\eqref{eq:Cupdate} using local approximation.
Specifically, let \[{ f({\bm c}[\ell];{\bm B}^t)=\frac{1}{2}\|{\bm x}[\ell]-{\bm B}^{t}{\bm c}[\ell]\|_2^2.}\]
Then, $f({\bm c}[\ell];{\bm B}^t)$ can be locally approximated at ${\bm c}^t[\ell]$ by the following:
{
\begin{align*}
 u({\bm c}[\ell];{\bm B}^t) &= f({\bm c}^t[\ell];{\bm B}^t)+\left(\nabla f({\bm c}^t[\ell];{\bm B}^t)\right)^T({\bm c}[\ell]-{\bm c}^t[\ell])\\
                   &\quad\quad\quad+ \frac{L^t}{2}\|{\bm c}[\ell]-{\bm c}^t[\ell]\|_2^2,
\end{align*}}
where $L^t\geq0$.
On the right hand side (RHS) of the above, the first two terms constitute a first-order approximation of { $f({\bm c}[\ell];{\bm B}^t)$} at ${\bm c}^t[\ell]$,
and the second term restrains ${\bm c}^{t+1}[\ell]$ to be close to ${\bm c}^t[\ell]$ in terms of Euclidean distance.
It is well-known that when $L^t \geq \|({\bm B}^t)^T{\bm B}^t\|_2$,
\[{ u({\bm c}[\ell];{\bm B}^t) \geq f({\bm c}[\ell];{\bm B}^t),~\forall {\bm c}[\ell]\in\mathbb{R}^K}\]
holds for all ${\bm c}[\ell]$ and the equality holds if and only if ${\bm c}[\ell]={\bm c}^t[\ell]$ \cite{hong2016unified}.
In other words, when $L^t \geq \|({\bm B}^t)^T{\bm B}^t\|_2$, { $u({\bm c}[\ell])$} is a `majorizing' function of $f({\bm c}[\ell];{\bm B}^t)$.
Given this majorizing function, we update ${\bm c}[\ell]$ by the following simple rule:
\begin{equation}\label{eq:approx_c}
	 {{\bm c}^{t+1}[\ell] =  \arg\min_{{{\bm 1}^T{\bm c}[\ell]=1,~{\bm c}[\ell]\geq {\bm 0}}}u({\bm c}[\ell];{\bm B}^t).}
\end{equation}
By re-arranging the terms and discarding constants, Problem~\eqref{eq:approx_c} is equivalent to the following
{
\begin{equation*}
\begin{aligned}
	          \min_{{{\bm 1}^T{\bm c}[\ell]=1,~{\bm c}[\ell]\geq {\bm 0}}} ~\left\|{\bm c}[\ell] -\left({\bm c}^t[\ell]- \frac{1}{L^t} \nabla f({\bm c}^t[\ell];{\bm B}^t)\right)\right\|_2^2.	
\end{aligned}
\end{equation*}
}
The RHS of the above can be considered as a gradient projection step with step size $1/{L^t}$.
Letting $P_{L^t}({\bm c}^t[\ell])$ denote the optimal solution of the above, we simplify the notation of updating ${\bm c}[\ell]$ as
\begin{equation}\label{eq:Cupper}
	  {\bm c}^{t+1}[\ell] =  P_{L^t}({\bm c}^t[\ell]).
\end{equation}
Problem~\eqref{eq:Cupper} is a simple projection that can be solved with worst-case complexity of ${\cal O}(K\log K)$ flops; see \cite{Wang2013} for a detailed implementation.

The described update of ${\bm C}$ has light per-iteration complexity,
but it could result in slow convergence of the overall alternating optimization algorithm; see Fig.~\ref{fig:converge} in the simulations.
To improve the convergence speed in practice, and inspired by the success of Nesterov's optimal first-order algorithm and 
its related algorithms
\cite{nesterov2004introductory,beck2009fast},
we propose the following update of ${\bm C}$:
\begin{subequations}\label{eq:C_FISTA}
\begin{align}
             {\bm c}^{t+1}[\ell]&=P_{L^t_c}({\bm y}^t[\ell])\\						
						       q^{t+1}&=\frac{1+\sqrt{1+4(q^t)^2}}{2}\\
                   {\bm y}^{t}[\ell] &= {\bm c}^t[\ell]+\left(\frac{q^{t}-1}{q^{t+1}}\right)\left({\bm c}^t[\ell]-{\bm c}^{t-1}[\ell]\right), \label{eq:proof}
\end{align}
\end{subequations}
where $\{q^t\}_{t=1}^\infty$ is a sequence with $q^1=1$.
Simply speaking,
instead of locally approximating $f({\bm c}[\ell];{\bm B}^t)$ at ${\bm c}^t[\ell]$,
we approximate it at an `extrapolated point' ${\bm y}^{t}[\ell]$.
Without the alternating optimization procedure, using extrapolation is provably much faster than using the plain gradient-based methods \cite{nesterov2004introductory,beck2009fast}.
Embedding extrapolation into alternating optimization was first considered in \cite{xu2013block} in the context of tensor factorization, where
acceleration of convergence was observed.
In our case, the extrapolation procedure also substantially reduces the number of iterations for achieving convergence, as will be shown in the simulations.


\subsection{Update of ${\bm B}$}
The update of ${\bm B}$ relies on the following two lemmas:
\begin{Lemma}\label{lem:conjugate}
    \cite{fu2015joint} Assume $0<p\leq 2$, $\epsilon> 0$, and let
		$\phi_p(w) := \frac{2-p}{2}\left(\frac{2}{p}w \right)^{\frac{p}{p-2}} +\epsilon w$.
		Then, we have
		$\left(x^2+\epsilon\right)^{p/2} = \min_{w\geq 0}~w x^2+\phi_p(w).$
		Also, the minimizer is unique and given by
    $w_{\rm opt} = \frac{p}{2}\left(x^2+\epsilon\right)^{\frac{p-2}{2}}$.
\end{Lemma}
\begin{Lemma}\label{lem:logdet}
\cite{jose2011robust} Let ${\bm E}\in\mathbb{R}^{K\times K}$ be any matrix such that ${\bm E}\succ {\bm 0}$.
Consider the function $f({\bm F})={\rm Tr}\left({\bm F}{\bm E}\right)-{\log}\det{\bm F}-K.$
Then, $\log\det {\bm E} = \min_{{\bm F}\succeq {\bm 0}}~ f({\bm F})$,
and the minimizer is uniquely given by
${\bm F}_{\rm opt} = {\bm E}^{-1}$.
\end{Lemma}
The lemmas provide two functions that majorize the data fitting part and the volume-regularization part in \eqref{eq:R_VolMin}, respectively.
Specifically, at iteration $t$ and after updating ${\bm C}$, we have $(\hat{\bm B}^t,\{{\bm c}^{t+1}[\ell]\}_{\ell=1}^L)$. Then, the following holds:
\begin{equation}\label{eq:B1}
	\log\det({\bm B}^T{\bm B}+\epsilon{\bm I}) \leq {\rm Tr}({\bm F}^t{\bm B}^T{\bm B})-{\log}\det{\bm F}^t-K,
\end{equation}
where ${\bm F}^t=(({\bm B}^t)^T{\bm B}^t+\epsilon{\bm I})^{-1}$ and the equality holds when ${\bm B}={\bm B}^t$.
Similarly, we have
\begin{equation}\label{eq:B2}
\begin{aligned}
	&\sum_{\ell=1}^L\frac{1}{2}\left(\left\|{\bm x}[\ell]-{\bm B}{\bm c}^{t+1}[\ell] \right\|_2^2+\epsilon\right)^{\frac{p}{2}} \\
	&\leq  \sum_{\ell=1}^L \frac{w^t_\ell}{2}\left\|{\bm x}[\ell]-{\bm B}{\bm c}^{t+1}[\ell] \right\|_2^2+\sum_{\ell=1}^L\phi_p(w^t_\ell),
\end{aligned}
\end{equation}
where $w_\ell^t = \frac{p}{2}(\|{\bm x}-{\bm B}^t{\bm c}^{t+1}[\ell]\|_2^2+\epsilon)^{\frac{p-2}{2}}$ and the equality holds when ${\bm B}={\bm B}^t$.
Putting \eqref{eq:B1}-\eqref{eq:B2} together and dropping the irrelevant terms, we find ${\bm B}^{t+1}$ by solving the following:
\begin{equation}\label{eq:Bupdate}
\begin{aligned}
	{\bm B}^{t+1}&:=\arg\min_{{\bm B}}\sum_{\ell=1}^L \frac{w_\ell}{2}\left\|{\bm x}[\ell]-{\bm B}{\bm c}^{t+1}[\ell] \right\|_2^2 \\
	             &\quad\quad\quad\quad+\frac{\lambda}{2} {\rm Tr}({\bm F}^t({\bm B}^T{\bm B})).
\end{aligned}
\end{equation}
Problem~\eqref{eq:Bupdate} is a convex quadratic program that admits
the following closed-form solution:
\begin{equation}\label{eq:Bsln}
\begin{aligned}
	 {\bm B}^{t+1} &:= {\bm X}{\bm W}^t({\bm C}^{t+1})^T\left({\bm C}^{t+1}{\bm W}({\bm C}^{t+1})^T+\lambda{\bm F}^t\right)^{-1},
\end{aligned}
\end{equation}
where ${\bm W}^t={\rm Diag}(w_1^t,\ldots,w_L^t)$.

\begin{Remark}
	{ The expression in \eqref{eq:Bupdate} reveals why the proposed criterion and algorithm can automatically downweight the effect brought by the outliers. Suppose that $({\bm B}^t,{\bm C}^{t+1})$ is a ``good enough'' solution which is close to the ground truth. Then, $w_\ell^t$ is small when ${\bm x}[\ell]$ is an outlier since the fitting error term $\|{\bm x}[\ell]-{\bm B}^t{\bm c}^{t+1}[\ell]\|_2^2$ is large. Hence, for the next iteration, ${\bm B}^{t+1}$ is estimated with the importance of the outlier ${\bm x}[\ell]$ downweighted.} 	
\end{Remark}

\begin{Remark}
In practice, adding constraints on ${\bm B}$ by letting ${\bm B}\in{\cal B}$ is sometimes instrumental, since a lot of applications do have prior information that can be used to enhance performance.
For example, in image processing, a nonnegative ${\bm B}$ is often sought, and thus one can set ${\cal B}=\mathbb{R}_{+}^{M\times N}$.
{ When ${\cal B}$ is convex, the problem in \eqref{eq:Bupdate} can usually be solved in an efficient manner; e.g., one can call general-purpose solvers such as interior-point methods.
However, using general-purpose solvers here may lose efficiency since solving constrained least squares to a certain accuracy {\em per se} may require a lot of iterations.}
To simplify the update, we update ${\bm B}^t$ following the same spirit of updating ${\bm C}$:
Let { $g({\bm B};{\bm C}^{t+1})=\sum_{\ell=1}^L \frac{w_\ell}{2}\left\|{\bm x}[\ell]-{\bm B}{\bm c}^{t+1}[\ell] \right\|_2^2 +\frac{\lambda}{2} {\rm Tr}({\bm F}^t({\bm B}^T{\bm B}))+{\rm const}$,
where ${\rm const}=\sum_{\ell=1}^L\phi_p(w^t_\ell)-K$.}
We solve a local approximation of $g({\bm B};{\bm C}^{t+1})$:
{
\begin{align}
	 {\bm B}^{t+1}&:=\arg\min_{{\bm B}\in{\cal B}} ~g({\bm B}^t;{\bm C}^{t+1})+\nabla g({\bm B}^t;{\bm C}^{t+1})^T({\bm B}-{\bm B}^t) \nonumber\\
	              &\quad\quad\quad\quad\quad\quad\quad\quad+\frac{\mu^t}{2}\|{\bm B}-{\bm B}^t\|_F^2 \nonumber\\
	&:= {\rm Proj}_{\cal B}\left({\bm B}^t - \mu^t \nabla g({\bm B}^t;{\bm C}^{t+1})\right), \label{eq:conB}
\end{align}}
where $\mu^t\geq 0$ and
{
\begin{align*}
\nabla g({\bm B}^t;{\bm C}^{t+1}) = &{\bm B}^t\left({\bm C}^{t+1}{\bm W}^t({\bm C}^{t+1})^T+\lambda{\bm F}^t\right) \\
                                            &- {\bm X}{\bm W}^t({\bm C}^{t+1})^T,
\end{align*}}
is the partial derivative of the cost function in \eqref{eq:Bupdate} w.r.t. ${\bm B}$ at ${\bm B}^t$,
and ${\rm Proj}_{\cal B}({\bm Z})$ denotes the Euclidean projection of ${\bm Z}$ on ${\cal B}$.
For some ${\cal B}$'s, the projection is easy to compute; e.g., when ${\cal B}=\mathbb{R}^{M}_{+}$,
we have
${\rm Proj}_{\cal B}\left({\bm Z}\right)=\max\{{\bm Z},{\bm 0}\};$
see other easily implementable projections in \cite{parikh2013proximal}.
Notice that the update in \eqref{eq:conB} can also easily incorporate extrapolation.
\end{Remark}

The robust volume minimization (RVolMin) algorithm is summarized in Algorithm~\ref{Algo:RVolMin-BSUM}.
Its convergence properties are stated in Proposition~\ref{prop:convergence}, whose proof is relegated to Appendix~\ref{app:prop}.
\begin{Prop}\label{prop:convergence}
{ Assume that $L^t$ and $\mu^t$ are chosen such that $L^t\geq\|({\bm B}^t)^T{\bm B}^t\|_2$ and $\mu^t\geq\|({\bm F}^t)^T{\bm F}^t\|_2$, respectively.
Also, assume that ${\cal B}$ is a convex closed set.
Then, if the initial objective value is finite, the whole solution sequence generated by Algorithm~\ref{Algo:RVolMin-BSUM} converges to the set $\cal S$ that consists of all the stationary points of Problem~\eqref{eq:R_VolMin}, i.e.,
\[\lim_{t\rightarrow \infty}~d^{(t)}\left( \left({\bm B}^t,{\bm C}^t\right),{\cal S}  \right) = 0,\]
where $d^{(t)}( ({\bm B}^t,{\bm C}^t),{\cal S}  ) =\min_{{\bm Y}\in {\cal S}}\|{\bm Y}-({\bm B}^t,{\bm C}^t)\|_F^2$.}
\end{Prop}

\begin{algorithm}[h]
\small
\SetKwInOut{Input}{input}
\SetKwInOut{Output}{output}
\SetKwRepeat{Repeat}{repeat}{until}

\Input{${\bm X}$; $p\in (0,2)$; $K$; initial $({\bm B},{\bm C})$; $\epsilon$; $\tau$. }

$t = 0$;

${\bm W}^t= {\bm I}$; ${\bm F}^t={\bm I}$; ${\bm B}^t = {\bm B}$; ${\bm C}^t = {\bm C}$;

\Repeat{Some stopping criterion is reached}{

	 select $L^t$ and $\mu^t$;

   ${\bm C}^{t+1}\leftarrow$ \eqref{eq:Cupper}, \eqref{eq:C_FISTA}, or ADMM in \cite{fu2016icassp}.

   ${\bm B}^{t+1}\leftarrow$ \eqref{eq:Bsln} or \eqref{eq:conB};
	
   $r \leftarrow r +1$;

$w_\ell^t \leftarrow \frac{p}{2}\left(\|{\bm x}[\ell]-{\bm B}^t{\bm c}^t[\ell]\|_2^2+\epsilon\right)^{\frac{p-2}{2}}$ for $\ell=1,\ldots,L$;

${\bm F}^t \leftarrow \left(({\bm B}^t)^T{\bm B}^t+\tau{\bm I}\right)^{-1}$.}

\Output{${\bm C}^t$; ${\bm B}^t$.}
\caption{RVolMin}\label{Algo:RVolMin-BSUM}
\end{algorithm}

\begin{Remark}
As mentioned before,
we may also use different volume regularizers.
Let us consider the volume regularizer in \eqref{eq:trace_vol} first.
It was shown in \cite{berman2004ice} that this regularizer can also be expressed as ${\rm vol}({\bm B})={\rm Tr}({\bm G}{\bm B}^T{\bm B})$, where ${\bm G}=K{\bm I}-{\bm 1}{\bm 1}^T$.
Therefore, by letting ${\bm F}^t={\bm G}$ in Algorithm~\ref{Algo:RVolMin-BSUM}, the updates can be directly applied to handle the regularizer in \eqref{eq:trace_vol}.
Dealing with \eqref{eq:vol_b} is more difficult.
One possible way is to make use of \eqref{eq:conB} since $\det({\bm B}^T{\bm B})$ is differentiable.
The difficulty is that a global upper bound of the subproblem w.r.t. ${\bm B}$ may not exist.
Under such circumstances, \emph{sufficient decrease} at each iteration needs to be guaranteed for establishing convergence to a stationary point \cite{razaviyayn2013unified}.
In practice, the \emph{Armijo rule} is usually invoked to achieve this goal,
which in general is computationally more costly compared to the cases where $\mu^t$ can be determined in closed form.
\end{Remark}

\begin{Remark}
Problem~\eqref{eq:R_VolMin} is a nonconvex optimization problem.
Hence, a good starting point of RVolMin can help reach meaningful solutions quickly.
In practice, different initializations can be considered:

\noindent
$\bullet$~Existing VolMin algorithms. Many VolMin algorithms, such as the ones working in the reduced-dimension domain (e.g., the algorithms in \cite{bioucas2009variable,MVES}), exhibit
good efficiency.
The difficulty is that these algorithms are usually sensitive to the DR process in the presence of outliers.
Nevertheless, one can employ robust DR algorithms together with the algorithms in \cite{bioucas2009variable,MVES} as an initialization approach.
Nuclear norm-based algorithms \cite{candes2011robust} are viable options for robust DR, but are not suitable for large-scale problems because of the computational complexity.
Under such circumstances, one may adopt simple alternatives such as that proposed in \cite{nie2014optimal}.

\noindent
$\bullet$~Nonnegative matrix factorization.
If ${\bm A}$ is known to be nonnegative, any NMF algorithm can be employed as initialization.
In practice, dealing with NMF is arguably simpler relative to VolMin, and many efficient solvers for NMF exist -- see \cite{huang2014putting} for a survey.
Although NMF usually does not provide a satisfactory result on its own in cases where it cannot guarantee the identifiability of its factors, using the NMF-estimated factors to initialize the algorithms that provide identifiability guarantees can sometimes enhance the performance of the latter.


%
\end{Remark}

\section{Simulations}
In this section, we provide simulations to showcase the effectiveness of the proposed algorithm.
We generate the elements of ${\bm A}\in\mathbb{R}^{M\times K}$ from the uniform distribution between zero and one.
We generate $\bm s[\ell]$ on the unit simplex and with $\max_i s_i[\ell] \leq \gamma$, where $\frac{1}{K} \leq \gamma \leq 1$ is given.
We choose $\gamma= 0.85$, which results in a so-called `no-pure-pixel case' in the context of remote sensing and is known to be challenging to handle; see \cite{Ma2013,gillis2014and} for details.
Zero-mean white Gaussian noise is added to the generated data.
To model outliers, we define the outlier at data point $\ell$ as ${\bm o}[\ell]$ and let ${\cal A}\subseteq\{1,\ldots,L\}$ be the index set of outliers.
We assume that ${\bm o}[\ell]={\bm 0}$ if $\ell\notin {\cal A}$ and ${\bm x}[\ell]={\bm o}[\ell]$ otherwise.
We denote $N_o=|{\cal A}|$ as the total number of outliers.
Those active outliers are generated following the uniform distribution between zero and one, and are scaled to satisfy problem specifications.
For the proposed algorithm, we fix $p=0.5$, $\epsilon=10^{-12}$, and $\tau=10^{-8}$ unless otherwise specified.
We stop the proposed algorithm when the absolute change of the cost function is smaller than $10^{-5}$
or the number of iterations reaches 1000.

We define the signal-to-noise ratio (SNR) as
${\rm SNR}=10\log_{10}\left(\frac{\mathbb{E}\{\|{\bm A}{\bm s}[\ell]\|_2^2\}}{\mathbb{E}\{\|{\bm v}[\ell]\|_2^2\}}\right)$.
Also, to quantify the corruption caused by the outliers, we define the signal-to-outlier ratio (SOR) as
${\rm SOR}=10\log_{10}\left(\frac{\mathbb{E}\{\|{\bm A}{\bm s}[\ell]\|_2^2\}}{\mathbb{E}\{\|{\bm o}[\ell]\|_2^2\}}\right)$.
We use the mean-squared-error (MSE) of ${\bm A}$ as a measure of factorization performance, defined as
\[
{\rm MSE}=
\min_{  \bm{\pi} \in \Pi }
\frac{1}{K} \sum_{k=1}^K \left\| \frac{ {\bm a}_{k} }{ \| {\bm a}_{k} \|_2 } -   \frac{ \hat{\bm a}_{{\pi_k}} }{ \| \hat{\bm a}_{{\pi_k}} \|_2 }    \right\|_2^2, 
\]
where $\Pi$ is the set of all permutations of $\{ 1,2,\ldots,K \}$; and $\hat{{\bm a}}_{k}$ is the estimate of ${\bm a}_k$.

In this section, we use the SISAL algorithm proposed in \cite{bioucas2009variable} as a baseline.
SISAL is a state-of-art robust VolMin algorithm that takes outliers into account by solving
\[\min_{\tilde{\bm B},{\bm 1}^T{\bm C}={\bm 1}^T,\{\tilde{\bm x}[\ell]=\tilde{\bm B}{\bm c}[\ell]\}}~\log\det(\tilde{\bm B})+\eta\|{\bm C}\|_h,\]
where $\|\cdot\|_h=\sum_{\ell=1}^L\sum_{k=1}^K\max(-c_k[\ell],0)$ is
an element-wise
hinge function.
The intuition behind SISAL is to penalize the outliers whose ${\bm c}[\ell]$ has negative elements, but still allowing them to exist, thereby having some robustness to outliers.
The tuning parameter $\eta>0$ in SISAL controls the amount of outliers that are ``allowed'', and we test multiple $\eta$'s for SISAL in the simulations.
We run the original SISAL that uses SVD-based dimension reduction and the modified SISAL which uses the robust dimension reduction (RDR) algorithm in \cite{nie2014optimal}.
The latter is also used to initialize the proposed algorithm.

\begin{figure}
	\centering
	\includegraphics[width=7.9cm]{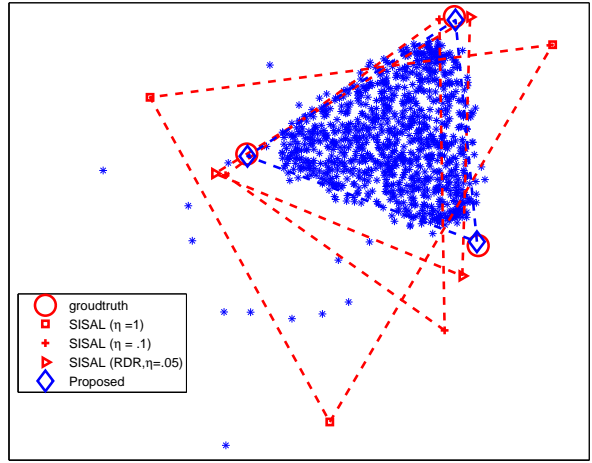}
	\caption{The $\hat{\bm A}$'s estimated by various algorithms. Blue points are ${\bm x}[\ell]$'s.}
	\label{fig:Toy}
\end{figure}

We first use an illustrative example to show the effectiveness of the proposed algorithm in the presence of outliers.
In this example, we set SNR$=18$dB, SOR$=-10$dB, $N_o=20$, $(M,K)=(50,3)$, and $L=1000$.
The results are projected onto the affine set that contains ${\rm conv}\{{\bm a}_1,{\bm a}_2,{\bm a}_3\}$, i.e., a two-dimensional hyperplane.
In Fig.~\ref{fig:Toy},
we see that SISAL with different $\eta$'s cannot yield reasonable estimates of ${\bm A}$ since the DR stage threw the data to a badly estimated subspace.
Using RDR, SISAL performs better, but is still not satisfactory.
In this case, the proposed algorithm yields the most accurate estimate of ${\bm A}$.

\begin{figure}
	\centering
		\includegraphics[width=7.9cm]{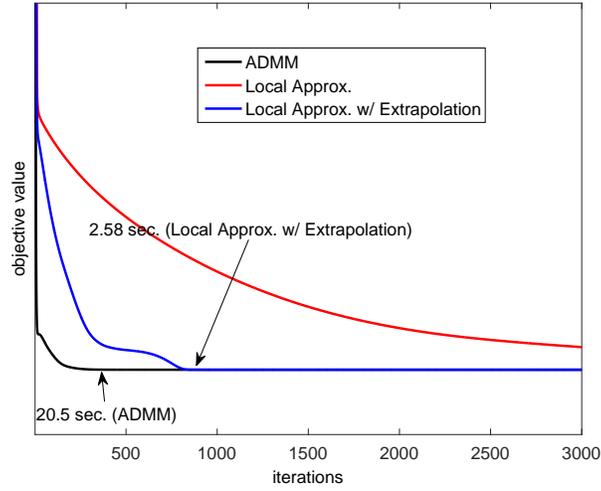}
  \caption{ Objective value vs. iterations, using different update strategies for ${\bm C}$.}
	\label{fig:converge}
\end{figure}

In Fig.~\ref{fig:converge}, we show the convergence curves of the algorithm under different update rules of ${\bm C}$, i.e., ADMM in \cite{fu2016icassp}, the proposed local approximation, and local approximation with extrapolation.
We show the results averaged from 10 trials, where SNR$=18$dB and SOR$=-5$dB.
We see that using ADMM, the objective value converges within 400 iterations.
Local approximation with extrapolation uses around 800 iterations to attain convergence of the objective value,
but the objective value cannot converge within 3000 iterations without extrapolation.
In terms of runtimes, the local approximation methods uses 0.003 second per iteration (a complete update of both ${\bm C}$ and ${\bm B}$),
while ADMM costs 0.05 second per iteration.
Obviously, local approximation with extrapolation is the most favorable update scheme:
its number of iterations for achieving convergence is around twice of that of ADMM,
but it is 15 times faster relative to ADMM for completing an update of ${\bm C}$.
{ Specifically, in the case under test, the average time for the algorithm using ADMM to update ${\bm C}$ to achieve the} { pointed objective value in Fig.~\ref{fig:converge} is 20.5 seconds, while using local approximation with extrapolation costs 2.58 seconds to reach the same objective level.}
In the upcoming simulations, all the results of the proposed algorithms are obtained with the extrapolation strategy.

\begin{figure}
	\centering
		\includegraphics[width=6cm]{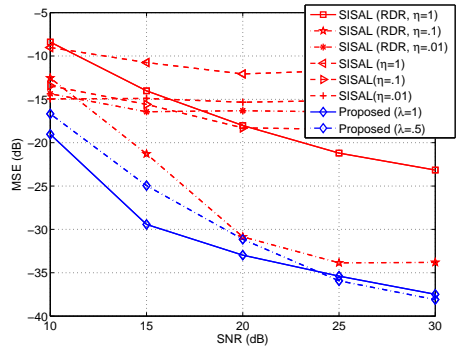}
	\caption{MSE of $\hat{\bm A}$ obtained by different algorithms vs. SNR. $(M,K)=(50,5)$; $N_o=20$; SOR$=-5$dB.}	
	\label{fig:SNRvary}
\end{figure}

In Fig.~\ref{fig:SNRvary}, we show the MSE performance of the proposed algorithm versus SNR.
We fix SOR$=-5$dB, and let $N_o=20$, $(M,K)=(50,5)$, and $L=1000$.
In Fig.~\ref{fig:SNRvary}, we see that
the original SISAL fails
for different $\eta$'s.
Using RDR, SISAL with $\eta=0.1$ yields reasonable results for all the tested SNRs.
The proposed algorithm with $\lambda=1$ and $\lambda=0.5$ gives the lowest MSEs.
The MSEs given by RVolMin with $\lambda=1$ are the lowest when SNR$\leq 20$dB,
and RVolMin with $\lambda=0.5$ exhibits the best MSE performance when SNR$\geq25$dB.
The results are consistent with the intuition behind selecting $\lambda$:
when
the
SNR is low, a relatively large $\lambda$ is needed to enhance the effect of the volume-minimization regularization.

\begin{figure}
	\centering
		\includegraphics[width=6cm]{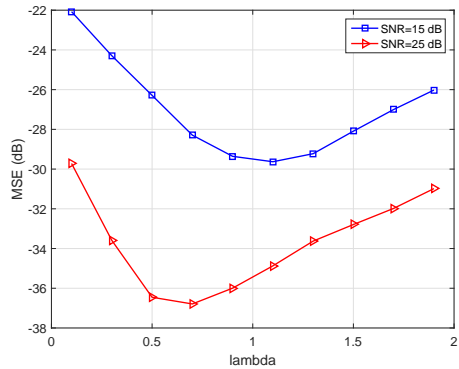}
	\caption{MSE of proposed algorithm vs. $\lambda$. $(M,K)=(50,5)$; $N_o=20$; $L=1000$; SOR$=-5$dB.}	
	\label{fig:lambda}
\end{figure}

To understand the effect of selecting $\lambda$, we plot the MSEs of the proposed algorithm versus $\lambda$ in Fig.~\ref{fig:lambda}.
We see that there exists an (SNR-dependent) optimal choice of $\lambda$ for achieving the lowest MSE,
but also note that any $\lambda$ in the range considered yields satisfactory results in both cases.


\begin{figure}
	\centering
    \includegraphics[width=6cm]{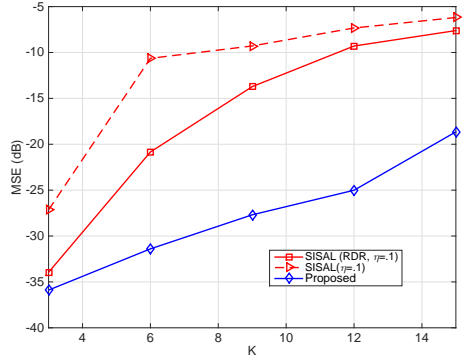}
	\caption{MSE of $\hat{\bm A}$ vs. $K$. $M=50$; $N_o=20$; $L=1000$; SOR$=-5$dB.}	
	\label{fig:Kvary}
\end{figure}

Fig.~\ref{fig:Kvary} shows the MSE performance
of the algorithms versus $K$.
We fix SNR$=20$dB and the other settings are the same as in the previous simulation.
The results of SISAL and SISAL with RDR are also used as baselines.
We run several $\eta$'s for SISAL and present the results of the one with the lowest MSEs.
{ As expected, all the algorithms work better when the rank of the factorization model is lower -- which is consistent with past experience on different matrix factorization algorithms, such as \cite{huang2014putting}.}
SISAL and SISAL with RDR work reasonably when $K=3$, but deteriorate when $K\geq 6$.
On the other hand, even when $K=15$, the proposed algorithm still works well, giving the lowest MSE.

\begin{figure}
	\centering
	\includegraphics[width=6cm]{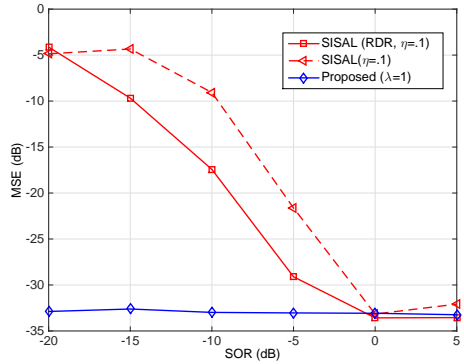}
	\caption{MSE of $\hat{\bm A}$ vs. SOR. $(M,K)=(50,5)$; $N_o=20$; $L=1000$; SNR$=20$dB.}	
	\label{fig:SORvary}
\end{figure}

Fig.~\ref{fig:SORvary} shows the MSEs of the algorithms versus SORs.
One can see that when some data are badly corrupted, i.e., when SOR$\leq -10$dB,
the proposed algorithm yields significantly lower MSEs than
SISAL and SISAL with RDR.
When SOR$\geq 0$dB,
all three algorithms provide comparable performance.

\begin{figure}
	\centering
		\includegraphics[width=6cm]{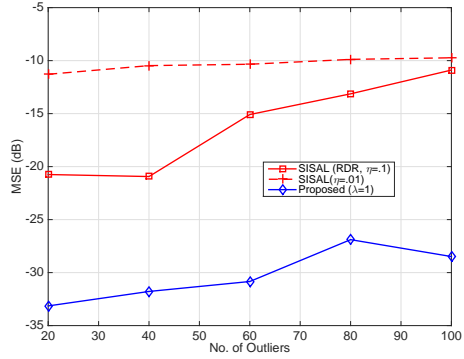}
	\caption{MSE of $\hat{\bm A}$ vs. $N_o$. $(M,K)=(50,5)$; $L=1000$; SOR$=-5$dB; SNR$=20$dB.}	
	\label{fig:Novary}
\end{figure}

We also test the algorithms versus the number of outliers. In Fig.~\ref{fig:Novary},
one can see that the proposed algorithm is not very sensitive to the change of $N_o$:
the MSE curve of the proposed algorithm is quite flat for different $N_o$'s in this simulation.
SISAL with RDR yields reasonable MSEs when $N_o\leq 40$, but its performance deteriorates when $N_o$ is larger.

\begin{table}[htbp]
  \centering
  \caption{The MSEs of the algorithms under ill-conditioned ${\bm A}$. $(M,K)=(50,5)$; $L=1000$; $N_o=20$; SOR$=-5$dB.}
  {
  \resizebox{8.5cm}{!}{\large
    \begin{tabular}{c|c|c|c|c}
    \hline
    \hline
    \multirow{2}[4]{*}{Algorithm} & \multicolumn{2}{c|}{uniformly distributed ${\bm A}$} & \multicolumn{2}{c}{ill-conditioned ${\bm A}$}\\
\cline{2-5}          & SNR=25dB & SNR=35dB & SNR=25dB & SNR=35dB\\
    \hline
    \hline
    SISAL & -12.6327 & -11.867 & -11.8829 & -11.8014\\
    \hline
    SISAL (RDR) & -22.3279 & -24.7461 & -13.2692 & -13.3246\\
    \hline
    Proposed ($\lambda=1$) & -35.5298 & -39.7004 & -24.6971 & {\bf -25.435}\\
    \hline
    Proposed ($\lambda=.5$) & {\bf -36.2388} & {\bf-41.5057} & {\bf -25.0232} & -25.3715\\
    \hline
    \hline
    \end{tabular}}}%
  \label{tab:ill}%
\end{table}%

{ Table~\ref{tab:ill} presents the MSEs of the estimated $\hat{\bm A}$ under well- and ill-conditioned ${\bm A}$'s, respectively. To generate an ill-conditioned ${\bm A}$, we use a way that is similar to the method suggested in \cite{Gillis2012}: in each trial, we first generate $\tilde{\bm A}$ whose columns are uniformly distributed between zero and one, and such $\tilde{\bm A}$'s are relatively well-conditioned. Then, we apply singular value decomposition to obtain $\tilde{\bm A}={\bm U}{\bm \Sigma}{\bm V}^T$. Finally, we replace ${\bm \Sigma}$ by $\tilde{\bm \Sigma}={\rm Diag}([1,0.1,0.01,0.005,0.001])$ and obtain ${\bm A}={\bm U}\tilde{\bm \Sigma}{\bm V}^T$. This way, the condition number of the generated ${\bm A}$ is $10^3$.
The other settings are the same as those in Fig.~\ref{fig:SNRvary}.
One can see from Table~\ref{tab:ill} that using such ill-conditioned ${\bm A}$, all the algorithms perform worse compared to the scenario where ${\bm A}$ has uniformly distributed columns (cf. the first and second columns in Table~\ref{tab:ill}). Nevertheless, the proposed algorithm still gives the lowest MSEs.}

In Table~\ref{tab:vol},
we present the MSE performance of the proposed algorithm using different volume regularizers.
We see that using ${\rm vol}({\bm B})={\rm Tr}({\bm G}{\bm B}{\bm B}^T)$ has the shortest runtime since the subproblem w.r.t. ${\bm B}$ is convex and can be solved in closed form.
When ${\rm vol}({\bm B})=\det({\bm B}^T{\bm B})$,
the algorithm requires much more time compared to that of the other two regularizers.
This is because the Armijo rule has to be implemented at each iteration.
In terms of accuracy, using the $\log\det({\bm B}^T{\bm B})$ regularizer gives the lowest MSEs when SOR$\leq -5$dB. Using $\det({\bm B}^T{\bm B})$ also exhibits good MSE performance when SOR$\geq 0$dB.
Using ${\rm Tr}({\bm G}{\bm B}{\bm B}^T)$ performs slightly worse in terms of MSE,
since it is a coarse approximation to simplex volume.
Interestingly, although our proposed log-determinant regularizer is not an exact measure of simplex volume as the determinant regularizer,
it yields lower MSEs relative to the latter.
Our understanding is that the performance gain results from the ease of computation.

Table~\ref{tab:nn} presents the MSE of the proposed algorithm with and without nonnegativity constraint on ${\bm B}$,
respectively.
We see that the MSEs are similar, with those of the nonnegativity-constrained algorithm being slightly lower.
This result validates the soundness of our update rule for the constrained case, i.e., \eqref{eq:conB}.
In terms of speed, the unconstrained algorithm requires less time.
We note that the nonnegativity constraint seems to only bring marginal performance gain in this simulation.
This might be because the data are generated following the model in \eqref{eq:basic_mod} and \eqref{eq:sumtoone}, and under this model VolMin identifiability does not depend on the nonnegativity of $\bm B$.
However, when we are dealing with real data, adding nonnegativity constraints makes much sense, as will be shown in the next section.

\begin{table}[htbp]
  \centering

  \caption{The MSEs of the proposed algorithm with different ${\rm vol}({\bm B})$'s. $(M,K)=(50,5)$; $L=1000$; $N_o=20$; SNR$=20$dB.}
     \resizebox{8.5cm}{!}{\large
    \begin{tabular}{c|c|c|c|c|c}
    \hline
    \hline
    \multirow{2}[4]{*}{${\rm vol}({\bm B})$} & \multirow{2}[4]{*}{measure} & \multicolumn{4}{c}{SOR (dB)} \\
\cline{3-6}          &       & -10   & -5    & 0     & 5 \\
    \hline
    \hline
    \multirow{2}[4]{*}{$\log\det({\bm B}^T{\bm B}+\tau{\bm I})$} & MSE   & {\bf -32.3289} & {\bf -33.1083} & {\bf -33.0075} & {\bf -32.9216} \\
\cline{2-6}          & TIME  & 6.262876 & 3.845999 & 3.328383 & 3.532759 \\
    \hline
    \multirow{2}[4]{*}{$\det({\bm B}^T{\bm B})$} & MSE   & -28.2461 & -24.0797 & -32.6881 & -32.1538 \\
\cline{2-6}          & TIME  & 91.0616 & 35.11194 & 38.98876 & 39.80187 \\
    \hline
    \multirow{2}[4]{*}{${\rm Tr}({\bm G}{\bm B}{\bm B}^T)$} & MSE   & -28.4332 & -28.5351 & -28.413 & -28.4776 \\
\cline{2-6}          & TIME  & {\bf 1.620391} & {\bf 1.540378} & {\bf 1.590791} & {\bf 1.517478} \\
    \hline
    \hline
    \end{tabular}%
}
  \label{tab:vol}%

\end{table}%

\begin{table}[htbp]
  \centering
  \caption{The MSEs of the proposed algorithm with and without nonnegativity constraint on ${\bm B}$ under various SNRs. $(M,K)=(50,5)$; $L=1000$; $N_o=20$; SOR$=5$dB.}
	\resizebox{8.5cm}{!}{\large
    \begin{tabular}{c|c|c|c|c|c}
    \hline
    \hline
    \multirow{2}[4]{*}{Algorithm} & \multirow{2}[4]{*}{measure} & \multicolumn{4}{c}{SNR (dB)} \\
\cline{3-6}          &       & 10    & 14    & 18    & 22 \\
    \hline
    \hline
    \multirow{2}[4]{*}{Proposed} & MSE   & -19.2268 & -28.2876 & -31.7755 & -33.7787 \\
\cline{2-6}          & TIME  & \textbf{1.737251} & \textbf{2.033748} & \textbf{2.875241} & \textbf{4.152018} \\
    \hline
    \multirow{2}[4]{*}{Proposed w/ nn} & MSE   & \textbf{-19.6422} & \textbf{-28.6822} & \textbf{-31.9502} & \textbf{-33.8563} \\
\cline{2-6}          & TIME  & 5.09489 & 5.943189 & 7.863959 & 11.10508 \\
    \hline
    \hline
    \end{tabular}%
		}
  \label{tab:nn}%
\end{table}%

\begin{figure}
	\centering
		\includegraphics[width=6cm]{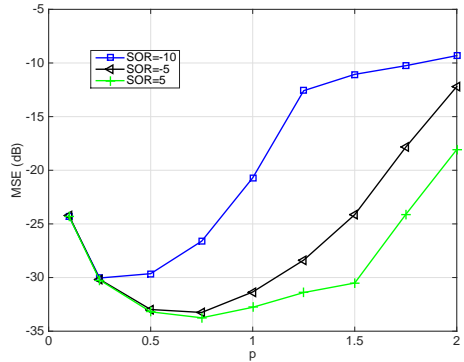}
		\caption{MSE of the proposed algorithm with different $p$'s under various SORs. $(M,K)=(50,5)$; $L=1000$; $N_o=20$; SNR$=20$dB.}	
	\label{fig:pvary}
\end{figure}

Fig.~\ref{fig:pvary} shows the effect of changing $p$.
When SOR$ = -10$dB, we see that using $p\in[0.25,0.75]$ gives relatively low MSEs.
This is because using a small $p$ is more effective in fending against outliers that largely deviate from the nominal model.
It is interesting to note that using $p=0.1$ gives slightly worse result compared to using $p\in[0.25,0.75]$.
Our understanding is that using a very small $p$ may
lead to
numerical problems, since the weights $\{{w}_\ell\}_{\ell=1}^L$
can be scaled in a very unbalanced way in such cases, resulting in ill-conditioned optimization subproblems.
For the cases where SOR$=-5$dB and $5$dB, a similar effect can be seen. In addition, a larger range of $p$, i.e., $p\in[0.25,1.5]$,
can result in good performance when SOR$=5$dB.
The results suggest a strategy of choosing $p$:
When the data is believed to be badly corrupt, using $p$ around $0.5$ is a good choice;
and when the data is only moderately corrupted,
using $p\in[1,1.5]$ is preferable,
since such a $p$ gives good performance and can better avoid numerical problems.

\section{Real Data Validation}
In this section, we validate the proposed algorithm using two real data sets,
i.e., a hyperspectral image dataset with known outliers and a document dataset.

\subsection{Hyperspectral Unmixing}
\begin{figure}
	\centering
		\includegraphics[width=6cm]{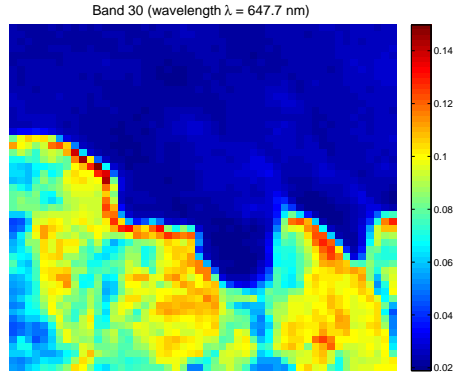}
	\caption{The considered subimage of the Moffet data set.}
	\label{fig:moffet}
\end{figure}

Hyperspectral unmixing (HU) is the application where VolMin-based factorization is most frequently applied;
see \cite{Ma2013}.
As introduced before, HU aims at estimating ${\bm A}$, i.e., the spectral signatures of the materials that are contained
in a hyperspectral image, and also their proportions ${\bm s}[\ell]$ in each pixel.
It is well-known that there are outliers in hyperspectral images, due to the complicated reflection environment, spectral band
contamination, and many other reasons \cite{dobigeon2014nonlinear}.
In this experiment, we apply the proposed algorithm to
a subimage of the real hyperspectral image that was captured over the Moffett Field in 1997 by the Airborne Visible/Infrared Imaging Spectrometer (AVIRIS) \footnote{Online available \url{http://aviris.jpl.nasa.gov/data/image_cube.html}}; see Fig.~\ref{fig:moffet}.
We remove the water absorption bands from the original 224 spectral bands, resulting in $M=200$ bands for each pixel ${\bm x}[\ell]$.
In this subimage with $50 \times 50$ pixels, there are three types of materials -- water, soil, and vegetation.
In the areas where different materials intersect, e.g., the lake shore, there are many outliers
as
identified by domain study.
Our goal here is to test whether our algorithm can identify the three materials and the outliers simultaneously.

\begin{figure}
	\centering
		\includegraphics[width=.49\textwidth]{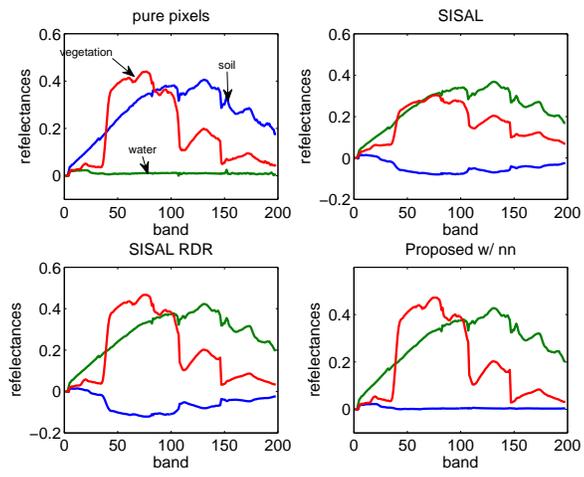}
	\caption{The spectra of the manually selected pure pixels and the estimated spectra by the algorithms.}
	\label{fig:signs}
\end{figure}

We apply SISAL, SISAL with RDR, and the proposed algorithm to estimate ${\bm A}$.
We set $\lambda=1$ for our algorithm and tune $\eta$ for SISAL carefully.
Notice that we let ${\cal B}={\mathbb{R}^{M\times K}_{+}}$ in this case, since the spectral signatures are known to be nonnegative.
The estimated spectral signatures are shown in Fig.~\ref{fig:signs}.
As a benchmark, we also present the spectra of some manually selected pixels, which are considered purely contributed by only one material,
and thus can approximately serve as the ground truth.
We see that both SISAL and SISAL with RDR
does not
yield accurate estimates of ${\bm A}$.
Particularly, the spectrum of water is badly estimated by both of these algorithms -- one can see
that there are many negative values of the spectra of water given by SISAL and SISAL with RDR.
On the other hand, the proposed algorithm with nonnegativity constraint on ${\bm B}$ gives spectra that are very similar to those of the pure pixels.

\begin{figure}
	\centering
		\includegraphics[width=.49\textwidth]{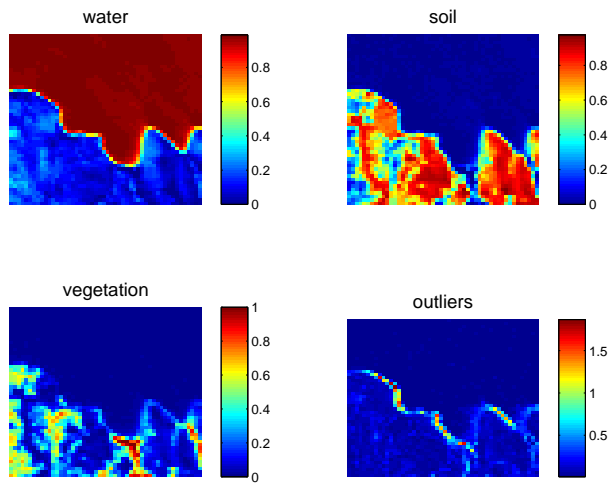}
	\caption{The abundance maps of the materials and the distribution of the outliers obtained by the proposed algorithm.}
	\label{fig:maps}
\end{figure}

Fig.~\ref{fig:maps} shows the spatial distributions of the materials (i.e., the abundance maps $\{{\bm s}_k[\ell]\}_{\ell=1}^L$ for $k=1,\ldots,K$)
that are estimated by the proposed algorithm in the first three subimages.
We see that the three materials are well separated,
and their abundance maps are consistent with previous domain studies \cite{fevotte2014nonlinear,halimi2011nonlinear}.
In the last subimage of Fig.~\ref{fig:maps},
we plot $1/{w}_\ell$ for $\ell=1,\ldots,L$.
Notice that the weight $w_\ell$ corresponds to the `importance' of the data ${\bm x}[\ell]$.
The algorithm is designed to automatically give small $w_\ell$ to outlying pixels.
We see that there are a lot of pixels on the lake shore that have very small weights, indicating that they are outliers.
Physically, these outliers correspond to those areas where the solar light reflects several times between water and soil,
resulting in nonlinearly mixed spectral signatures \cite{fevotte2014nonlinear,halimi2011nonlinear}.
The locations of the outlying pixels identified by our algorithm are also consistent with domain study \cite{fevotte2014nonlinear,halimi2011nonlinear}.

\subsection{Document Clustering}
We also present experimental results using the Reuters21578 document corpus\footnote{Online available: \url{http://www.daviddlewis.com/resources/testcollections/reuters21578/}}.
We use the subset of the full corpus provided by \cite{cai2011locally}, which contains 8,213 single-labeled documents from 41 clusters.
In our experiment, we test our algorithm under different $K$ (number of clusters), from 3 to 10.
Following standard pre-processing, each document is represented as a term-frequency-inverse-document-frequency (tf-idf) vector, and \emph{normalized cut weighting} is applied; see \cite{cai2011locally,xu2004document,manning2008introduction} for details.
We apply our VolMin algorithm to factor the document data ${\bm X}$ to `topics' ${\bm A}$ and a `weight matrix' ${\bm S}$ (cf.~Fig.~\ref{fig:motivation_rvolmin}),
and use ${\bm S}$ to indicate the cluster labels of the documents.
A regularized NMF-based approach, namely, \emph{locally consistent concept factorization} (LCCF) \cite{cai2011locally} is employed as the baseline,
which is considered a state-of-the-art algorithm for clustering the Reuters21578 corpus.
For each $K$, we perform 100 Monte-Carlo trials by randomly selecting $K$ clusters out of the total 41 clusters and 100 documents from each cluster.
We report the performance by comparing the results with the ground truth. Performance is measured by a commonly used metric called \emph{clustering accuracy}, whose detailed definition can be found in \cite{cai2011locally} -- the clustering accuracy ranges from 0 to 1, and higher accuracies indicate better performances.

Table~\ref{tab:doc} presents the results averaged from the 100 trials.
For the proposed algorithm, we set $\lambda=30$ and $p=1.5$; we also present the result of $p=2$ in this experiment, which we also use to initialize the $p<2$ case. Note that here we use a larger $\lambda$ relative to what was used in the simulations. The reason is that
the document corpus contains considerable modeling errors and the fitting residue is relatively large. The rule of thumb for selecting $\lambda$ is to set it at a similar level as the fitting error part; i.e., we let $\lambda = {\cal O}(\delta)$, where $\delta = \|{\bm X}-\hat{\bm A}\hat{\bm S}\|_F^2$ and can be coarsely estimated using plain NMF. This way, $\lambda$ can balance the fitting part and the volume regularizer. 
From Table~\ref{tab:doc}, we see that VolMin with $p=2$ already yields comparable clustering accuracy with LCCF. This indicates that, even without outlier-robustness, modeling the document clustering problem using VolMin-based SMF is effective.
Better accuracies can be seen by using $p=1.5$, where we see that for most $K$, the proposed RVolMin algorithm gives the best accuracy.
In particular, for $K=6,7$, more than $4\%$ accuracy improvement can be seen, which is considered significant in the context of document clustering.
Interestingly, further decreasing $p$
does not
yield better performance.
This implies that the modeling error is not very severe,
but outliers do exist, since using $p<2$ gives better clustering result than using $p=2$ which is not robust to modeling errors.

\section{Conclusion}
In this work, we looked into theoretical and practical aspects of the volume
minimization criterion for matrix factorization.
On the theory side, we showed that two independently developed sufficient conditions for VolMin identifiability are in fact equivalent.
On the practical side, we proposed an outlier-robust optimization surrogate of the VolMin criterion,
and devised an inexact BCD algorithm to deal with it.
Extensive simulations showed that the proposed algorithm outperforms a state-of-the-art robust VolMin algorithm, i.e., SISAL.
The proposed algorithm was also validated using real-world hyperspectral image data and document data, where
interesting and favorable results were observed.

\section*{Acknowledgment}
The authors would like to thank Prof. { Nicolas Dobigeon} for providing the subimage of the Moffet data.

\begin{table}[htbp]
  \centering
  \caption{The Clustering Accuracy on Reuters 21578 Corpus.}
    \begin{tabular}{c|c|c|c|c}
    \hline
    \hline
    \multirow{2}[4]{*}{algorithm} & \multicolumn{4}{c}{number of clusters} \\
\cline{2-5}          & 3     & 4     & 5     & 6 \\
    \hline
    LCCF  & 0.89042 & 0.8555 & 0.80367 & 0.73697 \\
    \hline
   Proposed ($p=2$) & 0.88387 & 0.83681 & 0.77565 & 0.74542 \\
    \hline
   Proposed ($p=1.5$)  & \textbf{0.92221} & \textbf{0.87376} & \textbf{0.81392} & \textbf{0.78672} \\
    \hline
    \hline
    \multirow{2}[4]{*}{algorithm} & \multicolumn{4}{c}{number of clusters} \\
\cline{2-5}          & 7     & 8     & 9     & 10 \\
    \hline
    LCCF  & 0.7131 & 0.65132 & \textbf{0.67441} & 0.6638 \\
    \hline
    Proposed ($p=2$)& 0.72719 & 0.65828 & 0.65244 & 0.64662 \\
    \hline
    Proposed ($p=1.5$)  & \textbf{{0.75217} }& \textbf{0.67852} & {0.67149} & \textbf{0.67708} \\
    \hline
    \hline
    \end{tabular}\label{tab:doc}
\end{table}%

\appendix

\ifplainver
    \section*{Appendix}
    \renewcommand{\thesubsection}{\Alph{subsection}}
\else
    \section{Appendix}
\fi

\subsection{Proof of Theorem~\ref{thm:equivalence}}\label{app:thm}

To show Theorem~\ref{thm:equivalence}, several properties of convex cones will be constantly used.
They are:
\begin{Property}\label{pro:dualcone}
	Let ${\cal K}_1$ and ${\cal K}_2$ be convex cones. Then,
	${\cal K}_1\subseteq {\cal K}_2 \Rightarrow {\cal K}_2^\ast\subseteq {\cal K}_1^\ast$.	
\end{Property}
\begin{Property}\label{pro:interseccone}
	Let ${\cal K}_1$ and ${\cal K}_2$ be convex cones. Then,
	$({\cal K}_1\cap {\cal K}_2)^\ast  = {\rm conv}\{{\cal K}_1^\ast \cup  {\cal K}_2^\ast\}$.	
\end{Property}
\begin{Property}\label{pro:unitcone}
	If ${\bm Q}$ is a unitary matrix. Then, ${\rm cone}({\bm Q})^\ast = {\rm cone}({\bm Q})$.
\end{Property}

We show Theorem~\ref{thm:equivalence} step by step.
First, we show the following lemma:
\begin{Lemma}\label{lem:conditions}
	Assume that ${\cal C}{\subseteq}{\rm cone}({\bm S})$ and ${\bm Q}$ is any unitary matrix except the permutation matrices. Then, we have
	${\rm cone}({\bm S})^\ast \cap {\rm bd}{\cal C}^\ast = \{\lambda_i{\bm e}_i | i=1,\ldots,N\}$$\Leftrightarrow$${\rm cone}({\bm S})\not\subseteq{\rm cone}({\bm Q})$.
\end{Lemma}

\emph{Proof}:
We first show the ``$\Rightarrow$'' part.
Given a unitary ${\bm Q}$, suppose that
\[{\rm cone}({\bm S})\subseteq {\rm cone}({\bm Q}).\]
By the basic properties of convex cones, we see that
${\rm cone}({\bm Q})^\ast \subseteq {\rm cone}({\bm S})^\ast$,
and ${\rm cone}({\bm Q})^\ast={\rm cone}({\bm Q})$.
Combining, we see
\begin{equation}\label{eq:QS}
{\rm cone}({\bm Q}) \subseteq {\rm cone}({\bm S})^\ast.
\end{equation}
Also, we have
\begin{equation}\label{eq:QC}
{\cal C}\subseteq{\rm cone}({\bm S})\Rightarrow {\rm cone}({\bm S})^\ast \subseteq {\cal C}^\ast\Rightarrow {\rm cone}({\bm Q})\subseteq {\cal C}^\ast.
\end{equation}
Combining Eq.~\eqref{eq:QS} and \eqref{eq:QC}, we have
\begin{equation}
{\rm cone}({\bm Q})\subseteq {\cal C}^\ast \cap {\rm cone}({\bm S})^\ast.
\end{equation}
We also know that the extreme rays of ${\rm cone}({\bm Q})$ lie in the boundary of ${\cal C}^\ast$, i.e., ${\rm ex}\{{\rm cone}({\bm Q})\}\subseteq {\rm bd}{\cal C}^\ast$ \cite[Lemma~1]{HuaSidSwa2014}. Thus, we have
\begin{equation}\label{eq:QSbdC}
{\rm ex}\{{\rm cone}({\bm Q})\}\subseteq {\rm bd}{\cal C}^\ast \cap {\rm cone}({\bm S})^\ast.
\end{equation}
Since we assumed ${\rm cone}({\bm S})^\ast \cap {\rm bd}{\cal C}^\ast = \{\lambda_i{\bm e}_i | i=1,\ldots,N\}$, we have
\begin{equation}
{\rm ex}\{{\rm cone}({\bm Q})\}\subseteq \{{\bm e}_1,\ldots,{\bm e}_N\}.
\end{equation}
Therefore, ${\bm Q}$ can only be a permutation matrix.

\bigskip

We now show the ``$\Leftarrow$'' part.
Following \eqref{eq:QSbdC}, and knowing that ${\rm cone}({\bm S})$ is a subset of the convex cone of some permutation matrix, we see that
\begin{equation}
\{{\bm e}_1,\ldots,{\bm e}_N\}\subseteq {\rm cone}({\bm S})^\ast\cap{\rm bd}{\cal C}^\ast.
\end{equation}
Now, suppose that there are a set of vectors $\{{\bm r}_1,\ldots,{\bm r}_p\}$ that does not include any unit vectors such that
\[{\rm cone}({\bm S})^\ast\cap{\rm bd}{\cal C}^\ast=\{{\bm e}_1,\ldots,{\bm e}_N,{\bm r}_1,\ldots,{\bm r}_p\}.\]
Then, we see that we can represent ${\rm cone}({\bm S})^\ast= {\rm conv}\{{\mathbb{R}_+^N}\cup {\rm cone}\{{\bm r}_1,\ldots,{\bm r}_p\}\}$.
By Property~\ref{pro:interseccone}, we see that
\begin{equation}\label{eq:Rr}
\begin{aligned}
{\rm cone}({\bm S})&={\rm conv}\{{\mathbb{R}_+^N}\cup {\rm cone}\{{\bm r}_1,\ldots,{\bm r}_p\}\}^\ast\\
                   &={\mathbb{R}_+^N}\cap {\rm cone}\{{\bm r}_1,\ldots,{\bm r}_p\}^\ast.
\end{aligned}
\end{equation}
Since ${\mathbb{R}_+^N}\cap {\rm cone}\{{\bm r}_1,\ldots,{\bm r}_p\}^\ast={\rm cone}({\bm S})\subseteq{\mathbb{R}^N_+}$, \eqref{eq:Rr} leads to
\[  {\rm cone}\{{\bm r}_1,\ldots,{\bm r}_p\}^\ast \subseteq \mathbb{R}_+^N. \]
By Property~\ref{pro:dualcone}, we see that
\[(\mathbb{R}_+^N)^\ast=\mathbb{R}_+^N\subseteq {\rm cone}\{{\bm r}_1,\ldots,{\bm r}_p\}.\]
This is a contradiction to the assumption that $\{{\bm r}_1,\ldots,{\bm r}_p\}$ does not include the unit vectors.
\hfill $\square$
\bigskip

Now we are ready to prove Theorem~\ref{thm:equivalence}.
We first notice that ${\cal C}\subseteq {\rm cone}({\bm S})$ is equivalent to $\gamma \geq \frac{1}{N-1}$.
In fact, we see that ${\cal C}\cap {\cal S}={\cal R}(\frac{1}{\sqrt{N-1}})$, where ${\cal S}=\{{\bm s}|{\bm 1}^T{\bm s}=1,{\bm s}\in\mathbb{R}^N\}$, and thus the claim holds.
Thus, our remaining work is to show that condition (ii) in Theorem~\ref{thm:FuMaHuaSid} is equivalent to restricting
$\gamma$ such that $\gamma>\frac{1}{\sqrt{N-1}}$

Step 1): Let us consider a conic representation of Theorem~\ref{thm:LinMa}.
Specifically, the corresponding convex cone of ${\cal R}(r)=\{{\bm s}\in\mathbb{R}^N|\|{\{{\bm s}\|_2\leq r\}} \cap {\rm conv}\{{\bm e}_1,\ldots,{\bm e}_N\}\}$  is $\tilde{\cal R}(r)={\cal C}(r)\cap {\mathbb{R}_+^N}$,
where
\[{\cal C}(r)=\{{\bm s}\in\mathbb{R}^N\mid \|{\bm s}\|_2\leq r{\bm 1}^T{\bm s}\},\]
and $\gamma={\rm sup}\{r|{\cal C}(r)\subseteq{\rm cone}({\bm S})\}$ under this definition.
It is also noticed that ${\cal C}(r)$ can be re-expressed as
\[{\cal C}(r)=\left\{ {\bm s} ~\middle|~ \frac{{\bm 1}^T{\bm s}}{\|{\bm 1}\|_2{\|{\bm s}\|_2}}\geq\frac{1}{r\sqrt{N}} \right\}.\]
In words, the vectors whose angles between ${\bm 1}$ are less than or equal to $\arccos\frac{1}{r\sqrt{N}}$ comprises ${\cal C}(r)$.
Therefore, by the definition of dual cone, i.e., ${\cal C}(r)^\ast=\{{\bm s}|{\bm s}^T{\bm y}\geq{\bm 0},~{\bm y}\in{\cal C}\}$, ${\cal C}(r)^\ast$ contains all the vectors that have the angle with ${\bm 1}$ less than or equal to $\arccos \frac{r}{\sqrt{r^2N-1}}$, which leads to
\begin{equation}\label{eq:calC}
{\cal C}(r)^\ast = {\cal C}\left(\frac{r}{\sqrt{r^2N-1}}\right).
\end{equation}

Step 2): Now, we consider the dual cone representation of Theorem~\ref{thm:LinMa}.
Let us define
\[{\cal T}(r)={\rm conv}({\cal C}(r)\cup{\mathbb{R}^N_+}). \]
Then, following Property~\ref{pro:interseccone} and \eqref{eq:calC}, we have
\begin{subequations}
	\begin{align}
	\tilde{\cal R}(r)^\ast  & =({\cal C}(r)\cap{\mathbb{R}^N_+})^\ast\\
	&=  {\rm conv}\left({\cal C}\left(\frac{r}{\sqrt{r^2N-1}}\right)\cup {\mathbb{R}_+^N}\right), \\
	& =  {\cal T}\left(\frac{r}{\sqrt{r^2N-1}}\right),
	\end{align}
\end{subequations}
where we have used $(\mathbb{R}_+^N)^\ast = \mathbb{R}_{+}^N$, i.e., Property~\ref{pro:unitcone}.
According to Property~\ref{pro:dualcone}, we see that
\begin{equation}\label{eq:coneST}
\tilde{\cal R}(r)\subseteq{\rm cone}({\bm S}) \Leftrightarrow {\rm cone}({\bm S})^\ast \subseteq {\cal T}\left(\frac{r}{\sqrt{r^2N-1}}\right).
\end{equation}
If we define $\kappa = {\rm inf}\{ r ~|~{\rm cone}({\bm S})^\ast \subseteq {\cal T}(r)  \}$,
we see from \eqref{eq:coneST} and the definition of $\gamma$ that
\begin{equation}
\gamma > \frac{1}{\sqrt{N-1}} \Leftrightarrow \kappa <1.
\end{equation}

Step 3): To show the equivalence between the sufficient conditions, let us begin from Theorem~\ref{thm:FuMaHuaSid}.
The condition ${\cal C}\subseteq {\rm cone}({\bm S})$ means that ${\cal C}^\ast\subseteq{\rm cone}({\bm S})^\ast$ by Property~\eqref{pro:dualcone}, and it further implies that $\{{\bm e}_1,\ldots,{\bm e}_N\}\subseteq{\rm ex}\{{\rm cone}({\bm S})^\ast\}$ \cite{HuaSidSwa2014}.
Suppose the rest of ${\rm cone}({\bm S})^\ast$'s extreme rays are ${\bm r}_1,\ldots,{\bm r}_p$, and $t$ is defined as
$t={\rm inf}\{r~|~{\rm cone}\{{\bm r}_1,\ldots,{\bm r}_p\}\subseteq{\cal C}(r) \}.$
Then, we have
\begin{align*}
{\rm cone}({\bm S})^\ast&={\rm cone}\{{\bm r}_1,\ldots,{\bm r}_p, {\bm e}_1,\ldots,{\bm e}_N\}\\
                        &\subseteq {\rm conv}\{{\cal C}(t)\cup {\mathbb{R}_+^N} \} ={\cal T}(t).
\end{align*}
Hence, by the definitions of $t$ and $\kappa$, we have $t= \kappa$.

Given the above analysis,
we first show that Theorem~\ref{thm:FuMaHuaSid} implies Theorem~\ref{thm:LinMa}.
Now, assume that Condition (ii) in Theorem~\ref{thm:FuMaHuaSid} is satisfied. We see that
${\rm cone}({\bm S})^\ast \cap {\rm bd}{\cal C}^\ast = \{\lambda_i{\bm e}_i | i=1,\ldots,N\}$ by Lemma~\ref{lem:conditions}.
Then, ${\bm r}_1,\ldots,{\bm r}_p$ are in the interior of ${\cal C}^\ast = {\cal C}(1)$. Therefore, we have $t<1$,
and subsequently $\kappa<1$ and $\gamma > \frac{1}{\sqrt{N-1}}$ strictly.

Now we show the converse by contradiction.
Assume that $\gamma> \frac{1}{\sqrt{N-1}}$ holds (the condition in Theorem~\ref{thm:LinMa} is satisfied).
If at least one point in ${\bm r}_1,\ldots,{\bm r}_p$ touches the boundary of ${\cal C}^\ast$ (condition (ii) in Theorem~\ref{thm:FuMaHuaSid} is not satisfied),
then $t=1$, and we cannot decrease it further while still contain ${\rm cone}({\bm S})^\ast$ in ${\cal T}(t)^\ast$.
This means that $\kappa = 1$, or, equivalently, $\gamma = \frac{1}{\sqrt{N-1}}$, which contradicts our first assumption that $\gamma> \frac{1}{\sqrt{N-1}}$.

\subsection{Proof of Proposition~\ref{prop:convergence}}\label{app:prop}
{ In the following, we prove the proposition under the algorithmic structure without extrapolation.
For the case where we update ${\bm C}$ using \eqref{eq:C_FISTA},
it is easy to see that ${\bm y}^t[\ell]\rightarrow{\bm c}^t[\ell]$ given $t \rightarrow \infty$ by \eqref{eq:proof}. Hence, if the proposition holds for the algorithm without extrapolation, it also holds for the extrapolated version asymptotically.

First, let us cast the proposed algorithm into the framework of \emph{block successive upper bound minimization} (BSUM) \cite{razaviyayn2013unified,hong2016unified}.
Unlike the classic block coordinate descent algorithm that solves every block subproblem exactly \cite{bertsekas1999nonlinear}, BSUM cyclically solves the upper-bound problems of every block subproblems.
We consider the updates using \eqref{eq:Cupper} and \eqref{eq:Bupdate} as an example.
The proof of using other updates will follow.
Our update rule in \eqref{eq:Cupper} and \eqref{eq:Bupdate} can be equivalently written as
\begin{subequations}
\begin{align}
{\bm C}^{t+1}&=\arg\min_{{\bm 1}^T{\bm C}={\bm 1}^T,{\bm C}\geq{\bm 0}}~u_C({\bm C};{\bm B}^{t}) \label{eq:ct}\\	
{\bm B}^{t+1}&=\arg\min_{\bm B}~u_B({\bm B};{\bm C}^{t+1}).
\end{align}
\end{subequations}
where $~u_C({\bm C};{\bm B}^{t})=\sum_{\ell=1}^L\frac{1}{2} (2u({\bm c}[\ell];{\bm B}^{t})+\epsilon)^{p/2}+\frac{\lambda}{2}\log\det(({\bm B}^t)^T{\bm B}^t+\tau{\bm I})$, $u({\bm c}[\ell];{\bm B}^{t})$ is defined as before, and
\begin{align*}
u_B({\bm B};{\bm C}^{t+1})=&\sum_{\ell=1}^L \frac{w_\ell}{2}\left\|{\bm x}[\ell]-{\bm B}{\bm c}^{t+1}[\ell] \right\|_2^2\\
&+\frac{\lambda}{2} {\rm Tr}({\bm F}^t({\bm B}^T{\bm B}))+{\rm const},
\end{align*}
in which ${\rm const}=\sum_{\ell=1}^L\phi_p(w^t_\ell)-K$.
Note that solving \eqref{eq:ct} is equivalent to solving \eqref{eq:Cupper} over different $\ell$'s since the problems w.r.t. $\ell=1,\ldots,L$ are not coupled.
Also denote $v({\bm B},{\bm C})$ as the objective value of Problem~\eqref{eq:R_VolMin}.
When $L^t\geq\|({\bm B}^t)^T{\bm B}^t\|_2$, we have
\begin{subequations}\label{eq:cond_1}
\begin{align}
v({\bm B}^t,{\bm C})&\leq u_C({\bm C};{\bm B}^{t}),\quad \forall {\bm C} \label{eq:C_ineq}\\
v({\bm B},{\bm C}^{t+1})&\leq u_B({\bm B};{\bm C}^{t+1}),\quad \forall {\bm B}, \label{eq:B_ineq}
\end{align}
\end{subequations}
where \eqref{eq:C_ineq} holds because under $L^t\geq\|({\bm B}^t)^T{\bm B}^t\|_2$ we have
\[f({\bm c}[\ell];{\bm B}^t)=\frac{1}{2}\|{\bm x}[\ell]-{\bm B}^t{\bm c}[\ell]\|_2^2\leq u({\bm c}[\ell];{\bm B}^t),~\forall {\bm c}[\ell],\]
and thus
\begin{align*}
 v({\bm B}^t,{\bm C}) & = \sum_{\ell=1}^{L}\frac{1}{2}\left( 2f({\bm c}[\ell];{\bm B}^t)+\epsilon \right)^{\frac{p}{2}}\\
                      &\quad+\frac{\lambda}{2}\log\det(({\bm B}^t)^T{\bm B}^t+\tau{\bm I})     \\
                          &\leq \sum_{\ell=1}^L \frac{1}{2}(2u({\bm c}[\ell];{\bm B}^{t})+\epsilon)^{\frac{p}{2}}\\
                          &\quad+\frac{\lambda}{2}\log\det(({\bm B}^t)^T{\bm B}^t+\tau{\bm I})\\
                          & = u_C({\bm C};{\bm B}^{t});
\end{align*}
Eq.~\eqref{eq:B_ineq} holds because of Lemmas~\ref{lem:conjugate} and \ref{lem:logdet};
also note that the equalities hold when ${\bm C}={\bm C}^t$ and ${\bm B}={\bm B}^t$, respectively.
Since all the functions above are continuously differentiable, we also have
\begin{subequations}\label{eq:cond_2}
\begin{align}
\nabla_{\bm C}v({\bm B}^t,{\bm C}^t)&=\nabla_{\bm C} u_C({\bm C}^t;{\bm B}^{t})\label{eq:grad_vc}\\
\nabla_{\bm B}v({\bm B}^t,{\bm C}^{t+1})&=\nabla_{\bm B} u_B({\bm B}^t;{\bm C}^{t+1}).\label{eq:grad_vb}
\end{align}
\end{subequations}
Note that if we update ${\bm C}$ using ADMM, we have
$v({\bm B}^t,{\bm C})= u_C({\bm C};{\bm B}^t)$ for all ${\bm C}$ -- Eqs.~\eqref{eq:C_ineq} and \eqref{eq:grad_vc} still hold.
Also, if we update ${\bm B}$ by \eqref{eq:conB}, the conditions in \eqref{eq:B_ineq} and \eqref{eq:grad_vb}
are also satisfied when $\mu^t\geq \|({\bm F}^t)^T{\bm F}^t\|_2$ since we now we have
\begin{align*}
u_B({\bm B};{\bm C}^{t+1})=& g({\bm B}^t;{\bm C}^{t+1})+\nabla g({\bm B}^t;{\bm C}^{t+1})^T({\bm B}-{\bm B}^t)\\
&+\frac{\mu_t}{2}\|{\bm B}-{\bm B}^t\|_F^2,
\end{align*}
and it can be shown that $v({\bm B},{\bm C}^{t+1}) \leq g({\bm B};{\bm C}^{t+1})\leq u_B({\bm B};{\bm C}^{t+1})$ and \[\nabla_{\bm B}v({\bm B}^t,{\bm C}^{t+1}) = \nabla_{\bm B} g({\bm B}^t;{\bm C}^{t+1}) = \nabla_{\bm B} u_B({\bm B}^t;{\bm C}^{t+1}), \]and the equalities hold simultaneously at ${\bm B}={\bm B}^{t}$.
Eqs~\eqref{eq:cond_1}-\eqref{eq:cond_2} satisfy the sufficient conditions for a generic BSUM algorithm to converge (cf. Assumption 2 in \cite{razaviyayn2013unified}). 
In addition, by \cite[Theorem~2 (b)]{razaviyayn2013unified},
if we can show that $({\bm B}^t,{\bm C}^t)$ lives in a compact set for all $t$, we can prove Proposition~\ref{prop:convergence}.}

{ Next, we show that in every iteration, ${\bm B}^t$ and ${\bm C}^t$ are bounded.
The boundness of ${\bm C}^t$ is evident because we enforce feasibility at each iteration. To show that ${\bm B}^t$ is bounded, we first note that
\[v({\bm B}^t,{\bm C}^t)\geq v({\bm B}^{t+1},{\bm C}^{t+1}),\]
where the inequality holds since the non-increasing property of the BSUM framework \cite{razaviyayn2013unified}.
By the assumption that $v({\bm B}^0,{\bm C}^0)$ is bounded, i.e.,
\[v({\bm B}^0,{\bm C}^0) \leq V,\]
where $V< \infty$, we have
\[\sum_{\ell=1}^L\frac{1}{2}\left(\left\|{\bm x}[\ell]-{\bm B}{\bm c}[\ell] \right\|_2^2+\epsilon\right)^{\frac{p}{2}} +\frac{\lambda}{2}  \log\det({\bm B}^T{\bm B}+\tau{\bm I})\leq V\]
holds for every $({\bm B},{\bm C})\in\{{\bm B}^t,{\bm C}^t\}_{t=1,\ldots,}$ that is generated by the algorithm. Since the first term on the left hand side of the above inequality is nonnegative, we have
\begin{subequations}
\begin{align}
&\log\det({\bm B}^T{\bm B}+\tau{\bm I})\leq V
\Leftrightarrow \log\left(\prod_{i=1}^{N}(\sigma_i^2+\tau)\right)\leq V \nonumber\\
&\Rightarrow  \log(\sigma_i^2+\tau) \leq V - (N-1)\log\tau,~\forall i  \label{eq:sigma1}\\
&\Rightarrow \sigma_i^2 \leq  \exp\left(V - (N-1)\log\tau\right) - \tau,~\forall i,  \label{eq:sigma2}
\end{align}
\end{subequations}
where $\sigma_1,\ldots,\sigma_N$ denote the singular values of ${\bm B}$,
and \eqref{eq:sigma1} holds since $\log(\sigma_i^2+\tau)\geq \log \tau$ for all $i$.
The right hand side of \eqref{eq:sigma2} is bounded, which implies that every singular value of ${\bm B}^t$ for $t=1,2,\ldots$ is bounded. Since ${\cal B}$ is a closed convex set, we conclude that the sequence $\{{\bm B}^t,{\bm C}^t\}_{t}$ lies in a compact set.
Now, invoking \cite[Theorem~2 (b)]{razaviyayn2013unified}, the proof is completed.}

\end{document}